%% file: mmcode_arr.tex
\documentclass[11pt]{article}

\usepackage[final]{acl}

\usepackage{times}
\usepackage{latexsym}

\usepackage[T1]{fontenc}
\usepackage[utf8]{inputenc}

\usepackage{microtype}

\usepackage{inconsolata}

\usepackage{graphicx}
\usepackage{wrapfig}
\usepackage{booktabs}

\usepackage{etoc}
\usepackage{tikz,graphics,color,fullpage,float,epsf,caption,subcaption,float,amsmath,multirow,booktabs,bbding,pifont,listings,changepage,tabularx,mdframed,arydshln,url}
\usepackage[frozencache,cachedir=./cache]{minted}

\usepackage[most]{tcolorbox}
\definecolor{green}{rgb}{0.345, 0.627, 0.333}

\definecolor{red}{rgb}{1, 0.25, 0.25}
\newcommand{\red}[1]{\textcolor{red}{#1}}

\title{
\includegraphics[width=0.6cm]{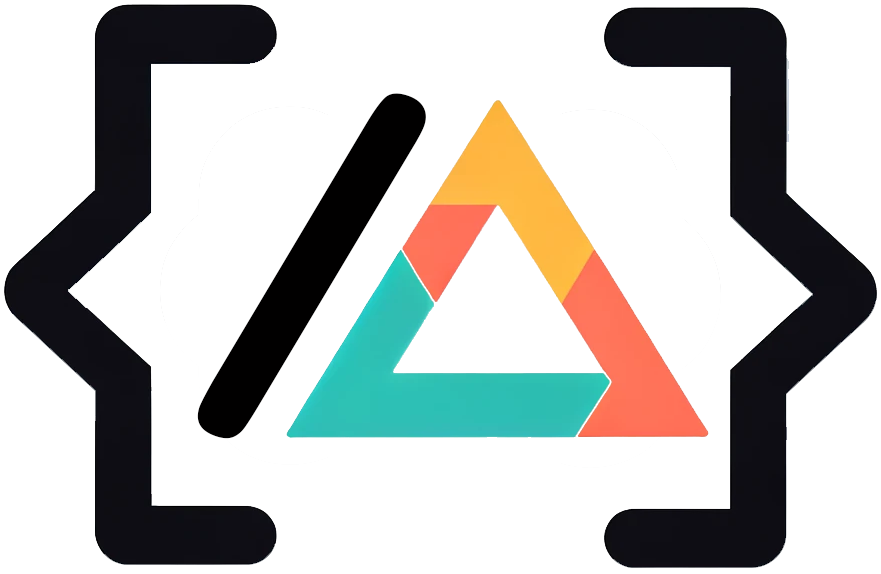}
MMCode: Benchmarking Multimodal Large Language Models in Code Generation with Visually Rich Programming Problems
}

\author{
Kaixin Li$^{1}$ \quad Yuchen Tian$^{2}$ \quad Qisheng Hu$^{3}$\ \quad Ziyang Luo$^{4}$ \\
\textbf{Zhiyong Huang$^{1}$}\thanks{~\hspace{1pt}Corresponding authors. } \quad \textbf{Jing Ma$^{4}$}\footnotemark[1]\\ 
  $^1$National University of Singapore \quad 
  $^2$The University of Hong Kong  \quad \\
  $^3$Nanyang Technological University \quad 
  $^4$Hong Kong Baptist University \\
 \texttt{likaixin@u.nus.edu}  \\ 
  }

\begin{document}
\maketitle
\begin{abstract}
Programming often involves converting detailed and complex specifications into code, a process during which developers typically utilize visual aids to more effectively convey concepts.
While recent developments in Large Multimodal Models have demonstrated remarkable abilities in visual reasoning and mathematical tasks, there is little work on investigating whether these models can effectively interpret visual elements for code generation. 
To this end, we present MMCode, the first multi-modal coding dataset for evaluating algorithmic problem-solving skills in visually rich contexts. MMCode contains 3,548 questions and 6,620 images collected from real-world programming challenges harvested from 10 code competition websites, presenting significant challenges due to the extreme demand for reasoning abilities.
% It presents significant challenges due to the extreme demand for reasoning abilities, the interwoven nature of textual and visual contents, coupled with the occurrence of questions containing multiple images. 
Our experiment results show that current state-of-the-art models struggle to solve these problems. 
The results highlight the lack of powerful vision-code models, and we hope MMCode can serve as an inspiration for future works in this domain. The data\footnote{\url{https://huggingface.co/datasets/likaixin/MMCode}} and code\footnote{\url{https://github.com/likaixin2000/MMCode}} are publicly available. 
% \footnote{\url{https://github.com/happylkx/MMCode}}

\end{abstract}

% Figure 1
\begin{figure*}[!t]
    \centering
    \includegraphics[width=0.95\textwidth]{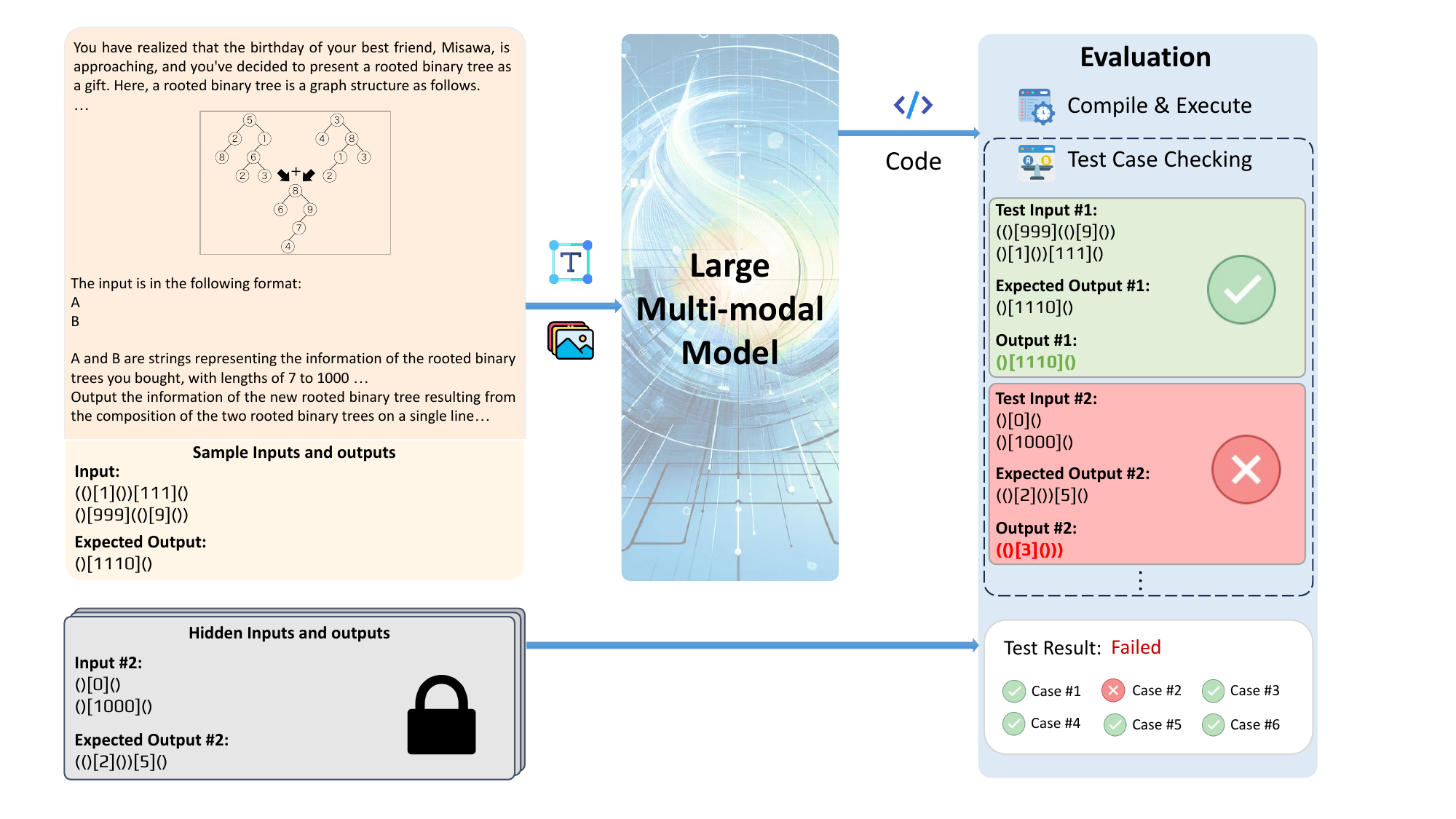}
    \caption{An illustration of an example question and the automatic testing pipeline of MMCode. The tests in the judger are selected for display. The actual test cases are harder than the sample inputs and outputs.}
    \label{fig:example}
\end{figure*}

\section{Introduction}
Programming is primarily aimed at fulfilling requirements, frequently entailing the translation of detailed and intricate specifications into executable code~\citep{nuseibeh2000software_requirements}. In this endeavor, human developers regularly employ visual aids such as images and diagrams to facilitate effective communication and a better understanding of concepts~\citep{agarwal2003obj_uml}.

Recently, automated code generation tools have attracted significant attention, largely attributing to the substantial advance in Code Large Language Models (Code LLMs)~\citep{chen2021codex,nijkamp2023codegen,roziere2023codellama,luo2023wizardcoder,li2023starcoder,guo2024deepseekcoder}. These models demonstrated unprecedentedly remarkable coding abilities, potentially assist to enhance productivity, reduce human error and democratize coding skills. Nevertheless, these models are limited to processing text-only inputs, lacking the ability to interpret rich information presented through images.

In a closely related development, the field has also observed the emergence of many powerful Large Multimodal Models (LMMs), marked by GPT-4V~\citep{2023gpt4v} and Gemini~\citep{team2023gemini}, representing a significant step forward in bridging the modality of text and images. While there are multiple works evaluating these models in mathematical reasoning~\citep{lu2023mathvista}, perception and reasoning~\citep{liu2023mmbench} and instruction-following~\citep{ye2023mplug}, there is a notable gap in evaluating LMMs for code generation.

To this end, we present MMCode, the first multi-modal benchmark for rigorously evaluating the code generation ability of Large Multimodal Models. It comprises 3,548 questions with 6,620 images collated from 10 programming-related websites encompassing a broad spectrum of subjects, extending from fundamental coding concepts to the application of code for solving mathematical problems. The generated code is rigorously checked by test cases. The overall framework is illustrated in Figure \ref{fig:example}.

Our experiments revealed that current LMMs struggle significantly to solve the tasks in MMCode. The most powerful LMMs, GPT-4V and Gemini, scored unsatisfactory pass rates as low as 19.4\% and 5.0\%, potentially due to the requirement of intense reasoning on the text descriptions and images. Open-source LMMs~\citep{liu2024llava,bai2023qwenvl} yield negligible pass rates because of their inability to understand the abstract meaning of the images. The findings reveal a significant deficiency in current LMMs' ability to interpret and utilize multimodal information for code generation, highlighting an imperative need for further advancements in this area. We believe MMCode will serve as a pivotal benchmark for evaluating the forthcoming evolution of Code LMMs and inspire research in this area.

\input{tables/image_categories}
\section{Related Works}

\subsection{Code Large Language Models}
Large Language Models (LLMs) have experienced significant advancements in recent years, demonstrating remarkable progress in their capabilities and applications that were previously unattainable~\citep{ouyang2022InstructGPT,brown2020GPT3,2022ChatGPT,2023GPT4,touvron2023llama,touvron2023llama2,chowdhery2022palm,anil2023palm2,hoffmann2022chinchila,scao2022bloom}. Building on their increasing proficiency at understanding and generating human-like text, a set of specialized models known as Code LLMs have emerged, focusing specifically on programming code~\citep{chen2021codex,nijkamp2023codegen,roziere2023codellama,li2023starcoder,luo2023wizardcoder,guo2024deepseekcoder}. 
Trained on large corpora of code data, these models have acquired the capacity to comprehend programming contexts and generate syntactically correct and logically sound code snippets. 
% This has not only made them invaluable tools for experienced developers~\citep{dakhel2023eval_copilot}, enhancing productivity and efficiency in software development processes, but also bridging the communication gap of human users and tool use, leading to intelligent agents~\citep{yang2024code_wand}. 
However, a significant limitation of these tools is their inability to process image inputs, restricting their application to environments where interaction is solely text- or code-based. Such a deficiency precludes their use in scenarios requiring the interpretation of visual data.

\subsection{Coding Benchmarks}
Accompanying the rapid development of Code Large Language Models, numerous benchmarks and datasets have witnessed the astonishing advancements of Code LLMs. These benchmarks cover a wide area of code-related tasks, such as code completion~\citep{chen2021codex,zheng2023codegeex,austin2021mbpp,yan2023codescope}, editing~\citep{li-etal-2024-instructcoder,tian2024debugbench} and translation~\citep{yan2023codetransocean}. 
% However, the relative simplicity of these datasets limits their effectiveness in assessing the performance of Code LLMs within more complex scenarios.
Most relevant to our work, APPS~\citep{2021apps} and CodeContests~\citep{2022alphacode} leveraged coding problems from real-world practice and contest coding websites as benchmarks. Recently, TACO~\citep{2023taco} contributed a comprehensive collection of contest problems. However, it aims to cluster the problems by the programming skills needed (e.g. Dynamic Programming and Tree Algorithms), while MMCode focuses on image-augmented questions to assess the question-solving skills of multi-modal language models.

\subsection{Reasoning-Intense Visual Question Answering}
Several works have emerged to assess the reasoning capabilities of LMMs with visual contexts.
ScienceQA~\citep{lu2022scienceqa} consists of multimodal multiple-choice questions across scientific topics, designed to measure the multi-hop reasoning ability.
MMMU~\citep{yue2023mmmu} features college-level questions with multi-disciplinary subjects.
MathVista~\citep{lu2023mathvista} emphasizes mathematical problem-solving with multi-modal input, involving tasks that require diverse math reasoning skills.
OlympiadBench~\citep{he2024olympiadbench} offers a set of challenging Olympiad-level mathematics and physics contest questions.
PuzzleVQA~\citep{chia2024puzzlevqa} benchmarks LMMs on patterns in order to evaluate if the models' reasoning ability generalizes to abstract figures. Our work distinguishes itself by necessitating the generation of solution code of complex problems, which benchmarks LMMs for long-horizon reasoning.

% Figure: Data Analysis
\input{figures/stats_questions}

\section{\includegraphics[width=0.5cm]{figures/logo.png} MMCode}

% In this section, we introduce the source and collection pipeline of MMCode. Our dataset contains programming questions from 10 platforms for diverse purposes. The collection pipeline comprises four stages: 1) Raw data collection; 2) automatic filtering; 3) human filtering and 4) annotation. The process starts with crawling raw questions from coding platforms. 
% % In this stage, we also attempt to reuse metadata and tests in TACO in the case of missing information due to policy changes and data expiry. 
% The collected raw data are converted to a unified format for further automatic processing, which mainly serves to remove questions without images or with broken images. Afterward, we perform a human inspection on every remaining question to further remove questions of low quality. Lastly, we annotate the images to provide the possibility of deeper analysis. Such a pipeline guarantees the quality and diversity of the data collected for MMCode.

In this section, we introduce the source and collection pipeline of MMCode. The collection pipeline comprises four stages: 1) Raw data collection; 2) automatic filtering; 3) human filtering and 4) annotation. 
% The process starts with crawling raw questions from coding platforms. 
% % In this stage, we also attempt to reuse metadata and tests in TACO in the case of missing information due to policy changes and data expiry. 
% The collected raw data are converted to a unified format for further automatic processing, which mainly serves to remove questions without images or with broken images. Afterward, we perform a human inspection on every remaining question to further remove questions of low quality. Lastly, we annotate the images to provide the possibility of deeper analysis. Such a 
This pipeline to be introduced in the following sections guarantees the quality and diversity of the data collected for MMCode.

\subsection{Data Sources}

The questions of MMCode are collected from 10 coding platforms, including AtCoder, Aizu, CodeChef, CodeForces, CodeWars, Project Euler, Geeksforgeeks, HackerRank, Leetcode and Open Kattis. 
% The questions are solved by users around the world, thereby ensuring their quality. 
More information can be found in Appendix \ref{appendix:sec:unrelated_img}.

The data sources exhibit a wide range of characteristics and purposes, including competitions, job interviews, and tutorials, etc. Notably, Project Euler is distinguished by its collection of challenges that necessitate a combination of mathematical and computer programming skills to solve. As a result, MMCode benefits from the diversity of these sources, offering programming problems with varying difficulties, styles, and skill requirements.

\subsection{Data Collection Pipeline}
\paragraph{Raw Data Collection.} For each of the 10 platforms, distinct web crawlers were developed to retrieve the problem statements. The HTML elements were then converted to plain texts following unified rules to ensure cleanliness and readability. Furthermore, the metadata of these questions was collected conditionally on availability, e.g. problem name, time limit, and memory limit. It is noteworthy that we also included the raw HTML code in our dataset for further flexible use. 
% For example, listing elements in \texttt{<ol>} and \texttt{<ul>} were rewritten as markdown listings (starting with ``\texttt{-}") across the sites. Superfluous new lines ``\texttt{\textbackslash{}n}'' were deleted in Open Kattis. The same was applied to \texttt{<pre>} blocks in Aizu. For Japanese-only questions on Aizu, we leveraged GPT-4 to translate them into English, while keeping the image tags. The numbers of image tags before and after translation were checked to make sure the image tags were correctly inserted. 

If there were images (\texttt{<img>} tags) encapsulated within the statements, we saved them and converted them to PNG format. The tags were replaced with markdown tags to insert them in the text (e.g. \texttt{![image](1.png)}). It is essential to maintain the locations of the images in the text because a question may encompass multiple images, and the images can be closely related to the text sections around them. This practice ensures the cohesion and coherence of the contents, where visual and textual elements are harmoniously integrated for better understanding.

Due to the difficulty of obtaining the automated test cases as a result of the changes in platforms' designs and policies, we also reused the rich information from the TACO dataset~\citep{2023taco} where feasible. We matched the crawled questions with those existing in TACO by URLs\footnote{TACO comprises questions obtained from mixed sources including CodeContests, APPS, and by crawling the websites. As a result, not all questions are provided with URLs. We abandoned the questions without URL metadata to ensure the overall quality of our data.}. Specifically, we crawled all questions from the largest two data sources, CodeForces and Aizu, including problem statements and test cases. Additionally, we included a new platform Project Euler that is not present in previous datasets. For other platforms, we reused the data from TACO and fetched the question statements to add the images. 

An initial data analysis revealed that 18.8\% of the obtained questions contained images, corroborating our motivation for creating a multi-modal coding benchmark.

\input{figures/stats_images}

\paragraph{Automated Filtering.}
In this phase, our initial step involved excluding questions that do not include associated images. Subsequently, we applied various post-processing steps to ensure the quality of the data. We filtered questions with images unable to load using the PILLOW library\footnote{https://github.com/python-pillow/Pillow}.
% Upon inspection, we found that the predominant cause of these failures was the expiration of links to third-party image hosting services, a frequent occurrence in user-created questions on CodeWars. Rare cases involved animated images (i.e. GIF format). These questions were eliminated to guarantee the loadability of all images. 
Additionally, we converted PNGs with alpha channels to pure RGB format by painting the background to pure white, which is critical for discerning the texts on the images. This avoids distinct behaviors of different models interpreting the transparent color. Finally, a strict 5-gram similarity is conducted on every pair of question statements in the dataset to remove similar problems with a similarity score greater than 0.80. This process eliminated 33 questions from the dataset.

% Following the described operations, we also removed questions with no available automated test cases to alleviate the problem of false positives~\cite{2022alphacode}. 

\paragraph{Human Filtering.}
At this stage, a preliminary inspection of sampled questions was first conducted to scope the quality of the collected data. The primary source of noise was found to be teaser images that try to interest the readers but do not provide information or implications to help solve the questions. These images mostly originate from Open Kattis and CodeForces, consisting of photographs about the background of the question, anime screenshots, etc. An example is presented in Appendix \ref{appendix:sec:unrelated_img}, where the question is about developers' cooperation, but the image is a humorous comic about the daily work of programmers. We also spotted some mixture of website logos and UI elements in the images, probably due to mistakes of the question creators in typesetting. 

To address this problem, a convenient solution is to employ large LMMs such as GPT-4V and Gemini to determine if the image(s) are useful in addressing the question. Nonetheless, such a method may potentially introduce bias into the data. Therefore, we decided to opt for human labor to filter out these unrelated images. We manually examined every image in the dataset to remove the noisy ones. Note that when an image was deemed irrelevant but was not the sole image in the question, we exclusively removed this image and its corresponding markdown tag from the text. The question itself is only eliminated if there are no images remaining after this process. 

\paragraph{Annotation.}
\label{sec:image_annotation}
In this stage, we annotate the images in MMCode into distinct categories in order to facilitate a more detailed analysis of model performance across various types of images. The images were examined and discussed by expert human coders who have rich experience in solving coding contest problems.
Following this deliberation, the images were meticulously categorized into 12 types: 
\texttt{Linear Data Structure}, \texttt{Tree}, \texttt{Graph}, \texttt{2D Geometry}, \texttt{3D Geometry}, \texttt{Chessboard}, \texttt{Map}, \texttt{Patterns}, \texttt{Math}, \texttt{Table}, \texttt{Pseudocode}, and \texttt{Others}. Gemini Pro Vision is leveraged to generate the coarse labels. Detailed descriptions of the categories are listed in Appendix \ref{appendix:sec:category_defs}.

This detailed categorization facilitates a focused analysis on how different types of visual information are processed and interpreted by models, thereby potentially aiding in the identification and improvement of their abilities in coding contexts.

\subsection{Data Splits}
After performing the previous procedures, we acquired a dataset with 3,548 questions with 6,620 images. Considering the lengthy nature of the questions and additional tokens needed to represent the images, evaluating on the full dataset can be expensive. Following MathVista~\citep{lu2023mathvista}, a conscious decision was made to keep the test set small. As a result, we sampled 263 questions as the test set, and applied careful human inspection to correct the image category labels.

\subsection{Testing Pipeline}
An execution-based testing pipeline is adopted in MMCode for rigorous answer checking, following~\cite{2021apps,2023taco,chen2021codex}. As demonstrated in Figure \ref{fig:example}, the judger attempts to compile the code generated by models, followed by a timed execution in a sandbox. The programs' outputs are checked against the ground truth answers in the test cases, and the solution is judged as correct only if it passes all hidden test cases. 

% Main results table
\input{tables/acc_by_category}

\section{Data Analysis}
In this section, we undertake a comprehensive exploration of MMCode,  introducing its and statistical attributes to provide a nuanced understanding of MMCode.

\paragraph{Problem Length.}
The diversity of data sources incorporated into MMCode results in significant variance in problem length, as can be seen in Figure \ref{fig:data_stats:length}. The mean length of the questions reaches 2,256 characters, with the 25th, 50th, and 75th percentile at 1,516, 2,127, and 2,791.
This can be ascribed to the distinct style and difficulty of the questions presented in MMCode. 
Certain questions articulate the instructions succinctly and directly, whereas others elaborate on the contextual background of the problem in detail. 
% Furthermore, difficult questions tend to have longer descriptions as they need to clearly convey the requirements and constraints.

\paragraph{Image Count per Problem.}
A notable characteristic that differentiates MMCode from previous datasets is its inclusion of multiple images per question. On average, each question is associated with 1.87 images, with the 25th percentile having 1 image and the 75th percentile having 2 images. These figures are interleaved with the text contents, and the understanding of them frequently depends on their order, posing great difficulty to the models.

\paragraph{Image Position.}
As Figure \ref{fig:image_stats:positions} illustrates, the images in the problems of MMCode can appear at any position in the text, but concentrate at the tail. This is because many images are drawn to intuitively depict and explain sample inputs and outputs, which are mostly located at the end of the text.

\paragraph{Image Type.}

Figure \ref{fig:image_stats:tags} illustrates the portion of the categories of images following the classification criteria introduced in Section \ref{sec:image_annotation}. \texttt{Graph}, \texttt{Math} and \texttt{2D Geometry} form the majority comprising more than half of the dataset, taking up 20.9\% 17.9\%, and 15.3\% respectively. Miscellaneous images classified under \texttt{Others} account for roughly one-tenth of the dataset, representing a high level of heterogeneity. \texttt{Tree} follows up with 9.9\%. The remaining groups sum up to approximately a quarter, demonstrating the diversity of MMCode. 

\input{tables/results_captioning}
\input{tables/results_cot}

\section{Experiments}

In this section, we benchmark several Language-Only models and Vision-Language models with MMCode. A comparative analysis of the experimental results for these models is conducted, providing a thorough examination of their capabilities.

\subsection{Experimental Setup}
We evaluate the models by prompting with fixed templates (see Appendix \ref{sec:prompts}) using greedy decoding and extracting their generated codes, which are executed by the testing framework to check their correctness. Pass@1~\citep{chen2021codex} is reported. The following three setups are compared:
\paragraph{Language-Only Models. }
We evaluate several powerful and Language-Only models, including GPT-3.5~\citep{2022ChatGPT}, GPT-4~\cite{2023GPT4}, and Gemini Pro~\cite{team2023gemini}. The images in the problem statement are removed in this setup.
\paragraph{Large Multi-modal Models. }
Some popular LMMs are selected as testees on MMCode. This includes proprietary models such as Gemini Pro Vision~\cite{team2023gemini}, GPT-4V~\cite{2023gpt4v}. Additionally, open-source models such as the LLaVA series~\citep{liu2024llava} and QWEN-VL~\cite{bai2023qwenvl} are assessed to track the advancements of the more accessible LMMs. The first image in the problem is kept for models that are not trained to support multiple-image inputs, i.e. the LLaVA series. For fairer comparison, text-only inputs performance of these models are also reported whenever applicable.

\paragraph{Caption-augmented Models. }
We investigate whether the inclusion of captions can help the model better understand the image contexts. In our early experiments, the open-source models yielded inferior captions, frequently containing hallucinations and failing to interpret the abstract meaning of the images. Thus, we only benchmark the proprietary models. We explored two methods of leveraging the captions: \textbf{(a) Image Replacement}, where the image slots are replaced by the captions. \textbf{(b) Captioning Chain of Thought}, where we explicitly prompt the models to generate captions for the images first, and then work out the questions, resembling the Chain of Thought prompting~\citep{wei2022cot}.
\section{Evaluation Results}
\subsection{Results and Findings}

% Error analysis figure
\input{figures/error_analysis}

\paragraph{MMCode poses a great challenge to all models.} 
As Table \ref{tab:main_results} depicts, all models except for the GPT family scored a Pass@1 rate under 10\%, whereas the best of the models tested, GPT-4V, yielded a mere 19.4\% when equipped with all image contexts. Test case pass rates, as a fine-grained measure, show a similar trend in Tabel \ref{appendix:table:test_case_avg} in Appendix \ref{appendix:test_case_avg}. This renders MMCode a challenging benchmark for the development of coding LMMs.

\paragraph{Proprietary models take a huge lead on MMCode.} 
The GPTs yield superior results, leaving a huge gap between other models. Gemini Pro, though underperforms the GPTs, beats all tested open-source models. The open-source models generally demonstrate the inability to solve the questions with negligible pass rates of around 1\% and a majority of zeros in many categories. A plausible reason is that these open-source models are not trained on such reasoning-heavy code generation tasks nor to understand abstract diagrams. The coding ability is only inherited from the base LLMs, but can be impaired due to catastrophic forgetting~\citep{luo2023catastrophic-forgetting}.

\paragraph{Visual context helps, but requires advanced comprehending capability.} Interestingly, unlike previous works such as OlympiadBench~\citep{he2024olympiadbench} where the text-only inputs beat multi-modal inputs, the best performance of all experiments is produced by GPT-4V with vision contexts. The observation confirms that the images contain critical information that can be mined to assist problem-solving. However, Gemini Pro Vision often fails to leverage the hints from the images, and the performance drops compared with the language-only Gemini Pro.

\paragraph{GPT-4V performs better than GPT-4 counterparts on less visually-cluttered image types.}
Comparing GPT-4V with multi-modal input to text-only GPT-4 and GPT-4V on problems with different types of images, it is observed that improvements are achieved on simpler image types, e.g. \texttt{Linear Data Structure}, \texttt{Tree}, \texttt{2D}, and \texttt{Map}. On other visually cluttered categories such as \texttt{Graph}, \texttt{Chessboard} and \texttt{Patterns}, the addition of images hurts the performance. GPT-4V also produces worse results on \texttt{Others}, which consists of miscellaneous cases including complex annotations, which are challenging for the model to interpret.

\paragraph{Image replacement with generated captions helps, but Captioning CoT does not.} Table \ref{tab:prompting_captioning} and \ref{tab:prompting_cot} lists the results with the two caption prompting strategies. The vision models can generate informative captions (though often inaccurate; see case studies in Section \ref{sec:case_study:captions}), as the text-only models all improve from their caption-free settings using the Image Replacement strategy. However, interestingly, all LMMs prompted with Captioning Chain of Thought suffer a decline in the pass rates. A possible explanation is that the captions lengthen the context, while the images still remain in the context, causing trouble for the models to determine where to attend.

\subsection{Error Analysis}

To facilitate the understanding of the models' bottleneck in solving MMCode problems, an identical subset of 50 questions are randomly selected from the failure cases of GPT-4V and LLaVA-13B and reviewed. Figure \ref{fig:error_analysis} presents the results. The majority of errors arise in the wrong understanding of the problems, where executable codes are generated but with wrong results. GPT-4V produces fewer runtime errors than LLaVA-13B, including Access Errors (e.g. IndexError, KeyError), Type Errors (e.g. calling non-existing methods of an object), and Math Errors (e.g. ZeroDivisionError). Notably, LLaVA-13B makes many elementary mistakes such as wrong Input Parsing and NameError (e.g. usage of variables undefined or defined afterward). These errors prevent the programs from producing outputs that can be checked, resulting in a decrease in Problem Understanding errors.

\subsection{Case Study}
\subsubsection{Caption Quality}
\label{sec:case_study:captions}
Figures \ref{fig:appendix:caption:linear} to \ref{fig:appendix:caption:others} in Appendix \ref{appendix:sec:case_study_captions} showcase the captions generated by GPT-4V and Gemini Pro Vision of 12 images from different categories. Generally, GPT-4V generates more accurate and more insightful captions than Gemini Pro Vision. However, both models can hallucinate the images, especially on visually complex elements such as \texttt{Graph} (Figure \ref{fig:appendix:caption:graph}). On the easier image of a \texttt{Tree} with fewer nodes and edges, both models produce correct explanations (Figure \ref{fig:appendix:caption:tree}). 

\subsubsection{Code Quality}
\label{sec:case_study:code}
We examined solutions generated by GPT-4V in the section in Appendix \ref{appendix:sec:code_case_study}). Apart from complex logic errors and inefficient implementations (Section \ref{appendix:sec:code_case_study:tle_wa}, it still makes trivial mistakes, e.g. naming variables after built-in functions (Section \ref{appendix:sec:code_case_study:name_conflict}), reading inputs when the problem does not ask it to (Section \ref{appendix:sec:code_case_study:input_parsing}).

% \section{Future Works and Potential Usage}

% \paragraph{Developing powerful Large Multimodal Models.}

% \paragraph{Programming Education.}
% MMCode contains real-world examples of how examiners utilize visual aids to facilitate the communication of concepts. It provides an important resource for guidance in creating figures to better teach the students. 

\section{Conclusion}
In this paper, we present MMCode, the first multi-modal coding dataset for evaluating algorithmic problem-solving skills in image-text interwoven contexts. We benchmarked a range of state-of-the-art LLMs and LMMs on MMCode and provide a detailed analysis. Despite their advanced capabilities, these models demonstrate a significant challenge in leveraging visual contexts for code generation. We believe that MMCode will catalyze further research and innovation, paving the way for the creation of AI systems capable of handling sophisticated visual and textual reasoning in programming and beyond.

\section{Limitations}
This work currently focuses on Python solutions, following the approaches outlined in~\citet{2023taco,2021apps}. Although support for additional programming languages could be achieved through similar execution-based verification, configuring the necessary environments for these languages is a task for future endeavors.

\section{Acknowledgement}
This work is partially supported by National Natural Science Foundation of China Young Scientists Fund(No. 62206233) and Hong Kong RGC ECS (No. 22200722).

% Bibliography entries for the entire Anthology, followed by custom entries
%\bibliography{anthology,custom}
% Custom bibliography entries only
\bibliography{custom}

\newpage
\appendix
\input{sections/appendix}

\end{document}

%% file: tables/image_categories.tex
\begin{table*}[t!]
\centering
\resizebox{\textwidth}{!}{
\begin{tabular}{|c|c|c|c|c|c|}
\hline
\includegraphics[width=2cm, keepaspectratio]{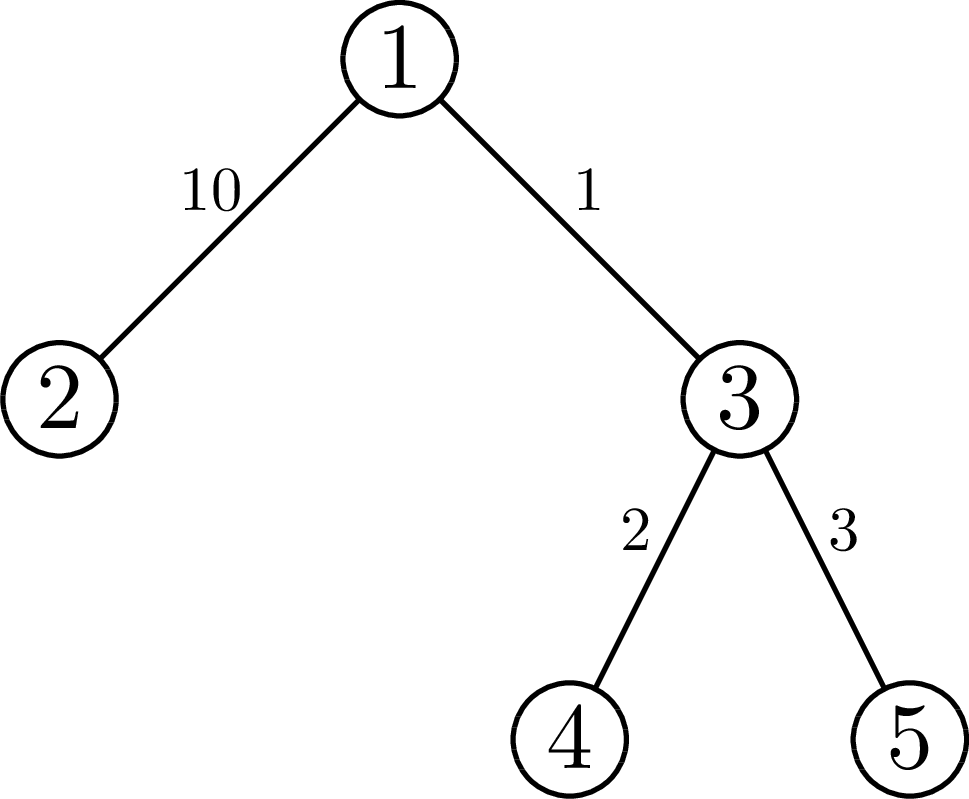} & 
\includegraphics[width=2cm, keepaspectratio]{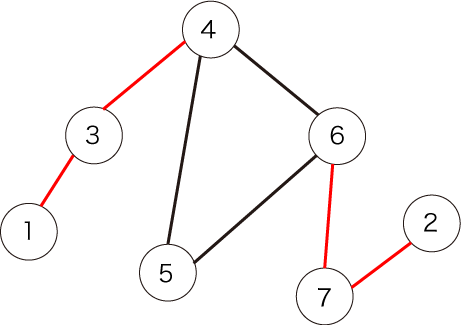} & 
\includegraphics[width=2cm, keepaspectratio]{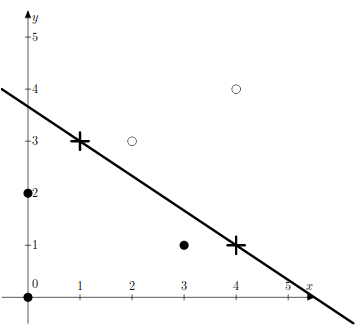} & 
\includegraphics[width=2cm, keepaspectratio]{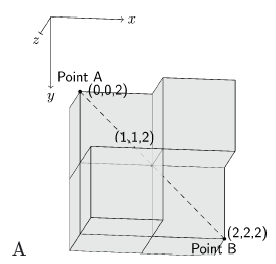} &
\includegraphics[width=2cm, keepaspectratio]{figures/image_categories/chessboard} &
\includegraphics[width=3cm, keepaspectratio]{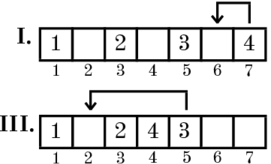} 
\\
 Tree & Graph & 2D & 3D & Chessboard & Linear Data Structure\\
\hline
\includegraphics[width=2cm, keepaspectratio]{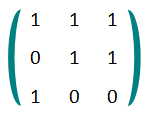} & 
\includegraphics[width=2cm, keepaspectratio]{figures/image_categories/patterns} & 
\includegraphics[width=2cm, keepaspectratio]{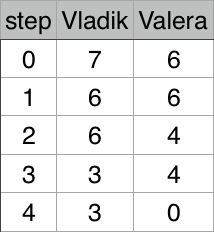} & 
\includegraphics[width=2cm, keepaspectratio]{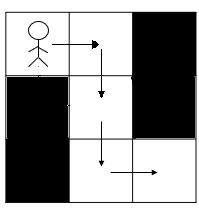} & 
\includegraphics[width=2cm, keepaspectratio]{figures/image_categories/others} &
\includegraphics[width=3cm, keepaspectratio]{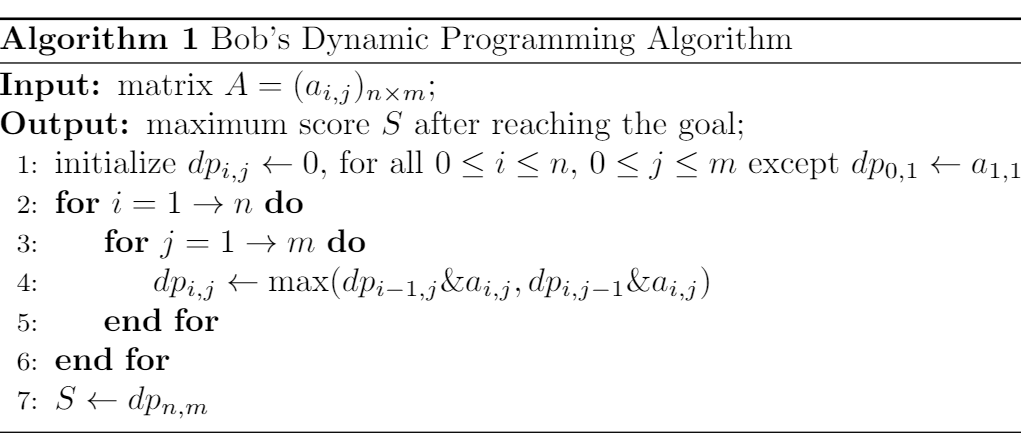}
\\
 Math & Pattern & Table & Map & Others & Pseudocode\\
\hline
\end{tabular}
}
\caption{Examples of images from each category. Some images are cropped for better visualization.}
\label{fig:image_categories}
\end{table*}

%% file: figures/stats_questions.tex
\begin{figure*}[t!]

% \begin{adjustwidth}{-1cm}{-1cm}

% \centering
\begin{subfigure}[t]{0.48\textwidth}
    \centering
    \includegraphics[width=1.\textwidth]{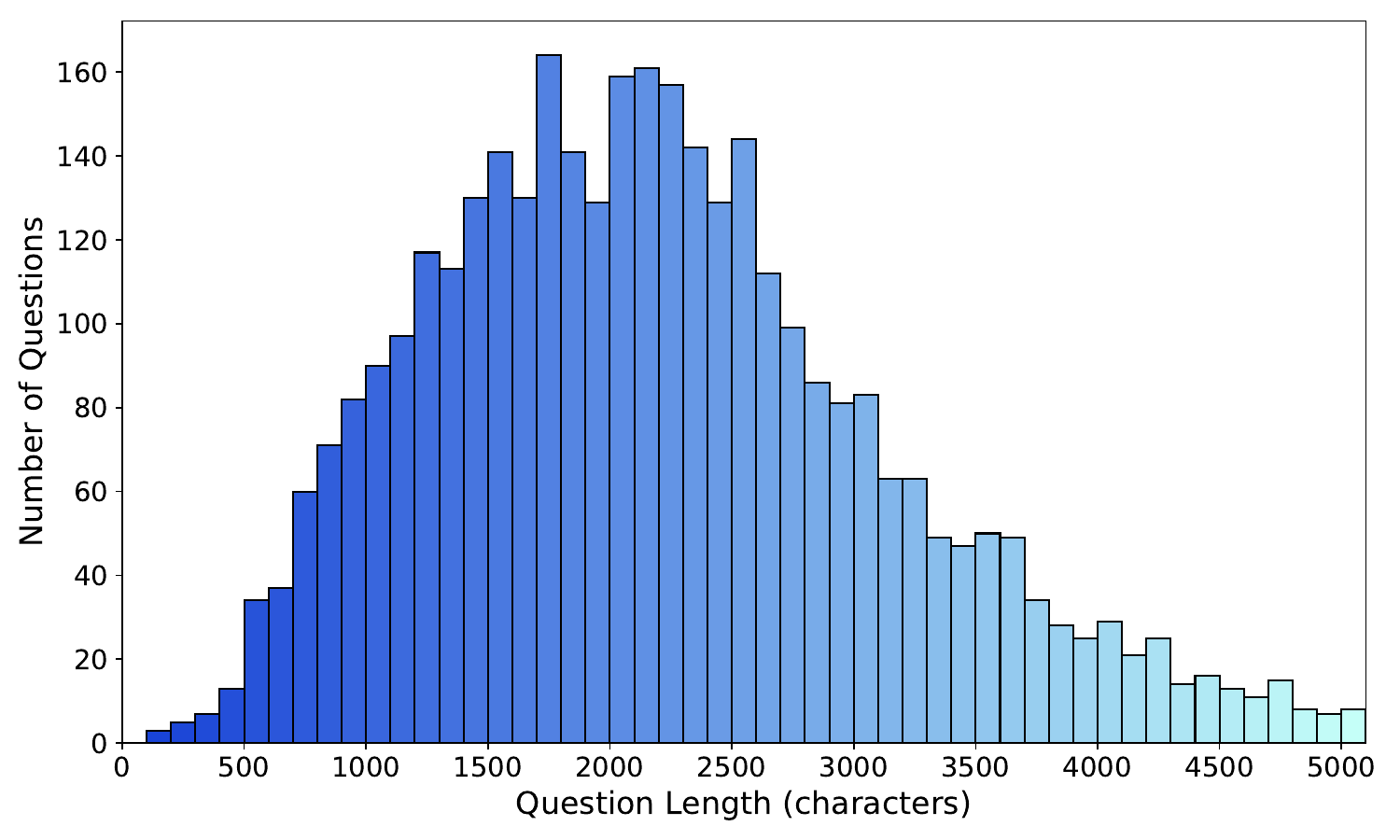}
    \caption{The distribution of lengths of the question statements, measured by the number of characters. }
    \label{fig:data_stats:length}
\end{subfigure}
\hfill
\begin{subfigure}[t]{0.48\textwidth}
    \centering
    \includegraphics[width=1.\textwidth]{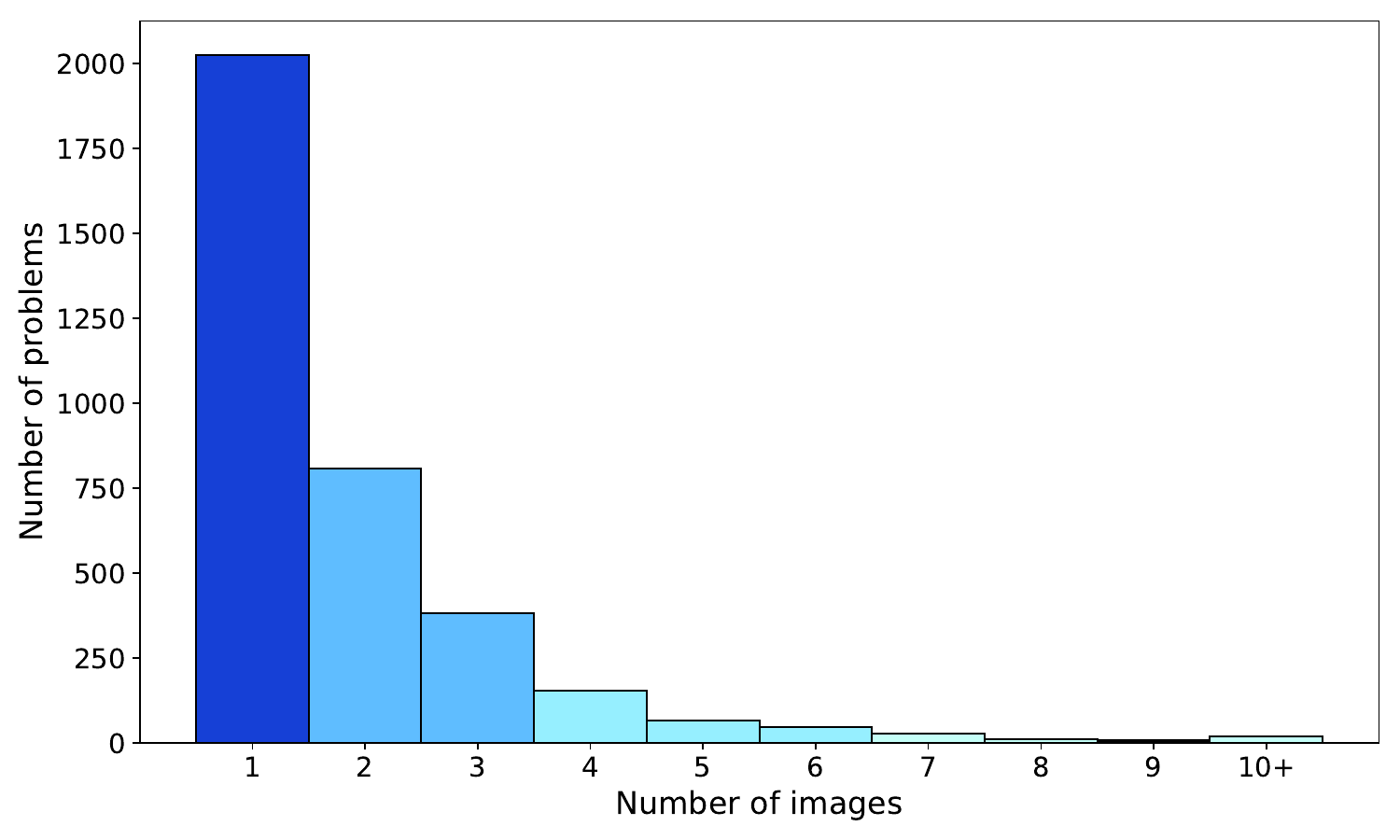}
    \caption{The distribution of numbers of images per question. Questions with 10 or more images are combined into the last bin ``10+''.}
    \label{fig:data_stats:image_count}
\end{subfigure}

     % \begin{subfigure}[b]{0.3\textwidth}
     %     \centering
     %     \includegraphics[width=\textwidth]{graph3}
     %     \caption{$y=5/x$}
     %     \label{fig:five over x}
     % \end{subfigure}
        \label{fig:three graphs}
\caption{Data statistics of the questions in MMCode. }
\label{fig:data_problem_stats}

% \end{adjustwidth}
    
\end{figure*}

%% file: figures/stats_images.tex
\begin{figure*}[t!]

% \begin{adjustwidth}{-1cm}{-1cm}

% \centering
\begin{subfigure}[t]{0.48\textwidth}
    \centering
    \includegraphics[width=1.\textwidth]{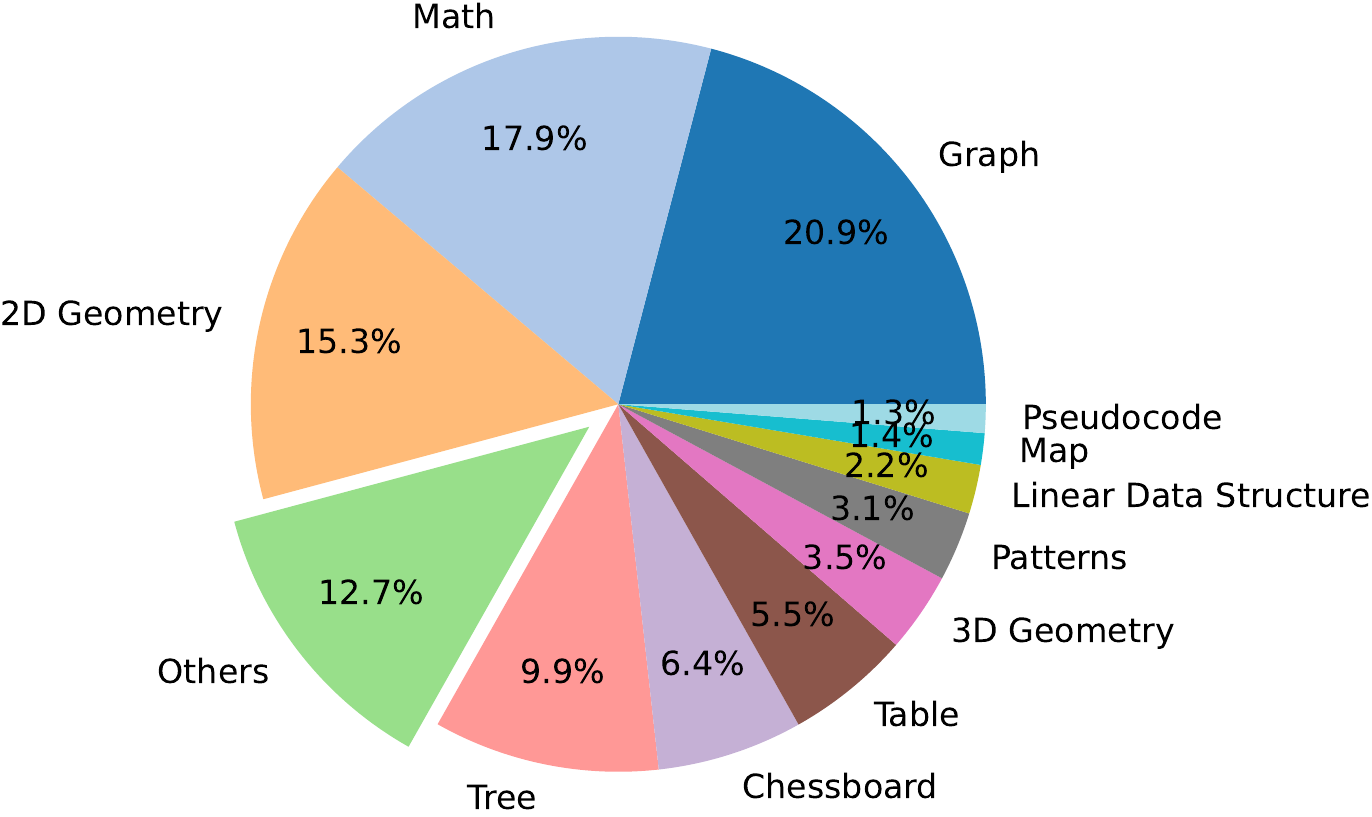}
    \caption{The distribution of lengths of the question statements, measured by the number of characters. }
    \label{fig:image_stats:tags}
\end{subfigure}
\hfill
\begin{subfigure}[t]{0.48\textwidth}
    \centering
    \includegraphics[width=1.\textwidth]{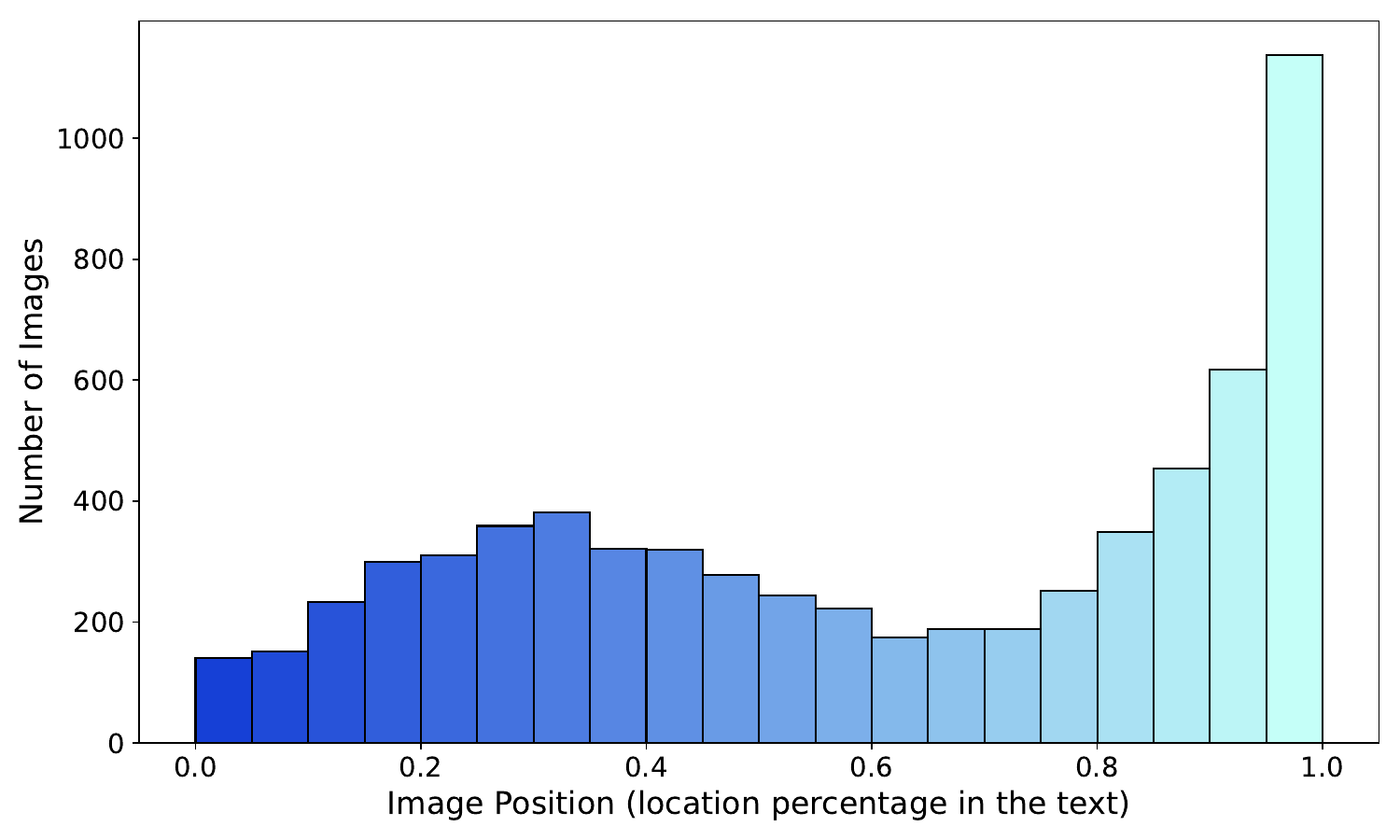}
    \caption{The distribution of the position of the images, measured by the percentage of their placement in the question statement. }
    \label{fig:image_stats:positions}
\end{subfigure}

     % \begin{subfigure}[b]{0.3\textwidth}
     %     \centering
     %     \includegraphics[width=\textwidth]{graph3}
     %     \caption{$y=5/x$}
     %     \label{fig:five over x}
     % \end{subfigure}
        
\caption{Data statistics of the images in MMCode. }
\label{fig:data_image_stats}

% \end{adjustwidth}
    
\end{figure*}

%% file: tables/acc_by_category.tex
\begin{table*}[t!]
\tiny
\centering
\tabcolsep 2pt
\resizebox{\textwidth}{!}{

\begin{tabular}{lccccccccccccc}
\toprule
\multirow{2}{*}{\textbf{Model}} & \multicolumn{12}{c}{\textbf{Task Type}} & \multirow{2}{*}{\textbf{Average}} \\ 
\cmidrule(lr){2-13}
& \textbf{Linear} & \textbf{Tree} & \textbf{Graph} & \textbf{2D} & \textbf{3D} & \textbf{Chessboard} & \textbf{Map} & \textbf{Math} & \textbf{Patterns} & \textbf{Table} & \textbf{Pseudocode} & \textbf{Others} &  \\ 
\midrule
\multicolumn{14}{c}{\textbf{Language Only Inputs}} \\
LLaVA-1.5-7B & 8.0 & 0.0 & 0.0 & 0.0 & 0.0 & 6.7 & 0.0 & 0.0 & 0.0 & 0.0 & 0.0 & 0.0 & 1.1\\
LLaVA-1.5-13B & 8.0 & 0.0 & 0.0 & 0.0 & 0.0 & 0.0 & 0.0 & 0.0 & 3.7 & 0.0 & 10.0 & 0.0 & 1.5\\
QWEN-VL & 4.0 & 0.0 & 0.0 & 0.0 & 0.0 & 6.7 & 0.0 & 0.0 & 0.0 & 0.0 & 10.0 & 0.0 & 1.1\\
CodeGemma-7b-Instruct & 12.0 & 0.0 & 0.0 & 0.0 & 3.8 & 6.7 & 3.6 & 0.0 & 3.7 & 0.0 & 20.0 & 0.0 & 3.4 \\
CodeLLaMA-7b-instruct & 8.0 & 0.0 & 0.0 & 0.0 & 0.0 & 0.0 & 0.0 & 0.0 & 0.0 & 0.0 & 10.0 & 0.0 & 1.1 \\
CodeLLaMA-13b-instruct & 8.0 & 0.0 & 0.0 & 0.0 & 0.0 & 6.7 & 0.0 & 0.0 & 0.0 & 0.0 & 10.0 & 0.0 & 1.5 \\
DeepSeekCoder-7b-instruct & 16.0 & 0.0 & 4.3 & 3.3 & 3.8 & 20.0 & 3.6 & 0.0 & 3.7 & 7.1 & 10.0 & 3.8 & 5.7 \\
DeepSeekCoder-33b-instruct & 16.0 & 0.0 & 8.7 & 10.0 & 7.7 & 20.0 & 17.9 & 8.0 & 11.1 & 7.1 & 30.0 & 11.5 & 11.4 \\
LLaMA3-instruct & 12.0 & 0.0 & 4.3 & 6.7 & 3.8 & 0.0 & 3.6 & 0.0 & 3.7 & 14.3 & 0.0 & 0.0 & 4.2 \\
MagiCoder-6.7b & 20.0 & 0.0 & 8.7 & 0.0 & 0.0 & 6.7 & 7.1 & 0.0 & 7.4 & 7.1 & 20.0 & 0.0 & 5.7 \\
StarCoder-15b-instruct & 12.0 & 0.0 & 0.0 & 6.7 & 0.0 & 0.0 & 0.0 & 4.0 & 7.4 & 0.0 & 10.0 & 0.0 & 3.4 \\
WizardCoder-15b & 8.0 & 0.0 & 0.0 & 3.3 & 0.0 & 6.7 & 0.0 & 0.0 & 3.7 & 0.0 & 20.0 & 0.0 & 2.7 \\
\hdashline
Gemini Pro & 16.0 & 0.0 & 4.3 & 3.3 & 0.0 & 0.0 & 3.6 & 0.0 & 14.8 & 0.0 & 20.0 & 7.7 & 5.7\\
GPT-3.5 \scalebox{0.7}{\texttt{(gpt-3.5-turbo-1106)}} & 28.0 & 6.9 & 4.3 & 6.7 & 7.7 & 13.3 & 10.7 & 4.0 & 18.5 & 14.3 & 20.0 & 7.7 & 11.0\\
GPT-4 \scalebox{0.7}{\texttt{(gpt-4-1106-preview)}} & 28.0 & 6.9 & 13.0 & 10.0 & 7.7 & 13.3 & 17.9 & 16.0 & 29.6 & 21.4 & 40.0 & 26.9 & 17.9\\
GPT-4V \scalebox{0.7}{\texttt{(gpt-4-vision-preview)}} & 40.0 & 10.3 & 17.4 & 10.0 & 7.7 & 26.7 & 7.1 & 12.0 & 22.2 & 21.4 & 50.0 & 23.1 & 18.3\\
GPT-4o \scalebox{0.7}{\texttt{(gpt-4o-2024-05-13)}}  & 32.0 & 6.9 & 8.7 & 3.3 & 11.5 & 20.0 & 10.7 & 16.0 & 18.5 & 7.1 & 40.0 & 15.4 & 14.8 \\
\midrule
\multicolumn{14}{c}{\textbf{Vision + Language Inputs}} \\
LLaVA-1.5-7B & 12.0 & 0.0 & 0.0 & 0.0 & 0.0 & 6.7 & 0.0 & 0.0 & 0.0 & 0.0 & 0.0 & 0.0 & 1.5\\
LLaVA-1.5-13B & 8.0 & 0.0 & 0.0 & 0.0 & 0.0 & 0.0 & 0.0 & 0.0 & 3.7 & 0.0 & 0.0 & 0.0 & 1.1\\
QWEN-VL & 8.0 & 0.0 & 0.0 & 0.0 & 0.0 & 0.0 & 0.0 & 0.0 & 0.0 & 0.0 & 0.0 & 0.0 & 0.8\\
\hdashline
Gemini Pro Vision & 12.5 & 0.0 & 4.3 & 0.0 & 3.8 & 6.7 & 7.1 & 0.0 & 7.4 & 0.0 & 30.0 & 0.0 & 5.0\\
GPT-4V \scalebox{0.7}{\texttt{(gpt-4-vision-preview)}} & 40.0 & 6.9 & 13.0 & 13.8 & 3.8 & 21.4 & 24.0 & 9.5 & 25.9 & 21.4 & 40.0 & 20.8 & 19.4\\
GPT-4o \scalebox{0.7}{\texttt{(gpt-4o-2024-05-13)}} & 36.0 & 6.9 & 8.7 & 3.4 & 7.7 & 21.4 & 24.0 & 14.3 & 25.9 & 14.3 & 50.0 & 8.3 & 17.0 \\
\bottomrule
\end{tabular}

}
\caption{Pass@1 (\%) results grouped by different image categories. The dashed lines separate open-source models (above) and proprietary models (below).}
\label{tab:main_results}
\end{table*}

%% file: tables/results_captioning.tex
\begin{table*}[t!]
\tiny
\centering
\tabcolsep 2pt
\resizebox{\textwidth}{!}{

\begin{tabular}{lccccccccccccc}
\toprule
\multirow{2}{*}{\textbf{Model}} & \multicolumn{12}{c}{\textbf{Task Type}} & \multirow{2}{*}{\textbf{Average}} \\ 
\cmidrule(lr){2-13}
& \textbf{Linear} & \textbf{Tree} & \textbf{Graph} & \textbf{2D} & \textbf{3D} & \textbf{Chessboard} & \textbf{Map} & \textbf{Math} & \textbf{Patterns} & \textbf{Table} & \textbf{Pseudocode} & \textbf{Others} &  \\ 
\midrule
Gemini Pro & 16.0 & 0.0 & 0.0 & 6.7 & 0.0 & 6.7 & 3.6 & 0.0 & 11.1 & 7.1 & 20.0 & 7.7 & 6.1\\
GPT-4 \scalebox{0.7}{\texttt{(gpt-4-vision-preview)}} & 32.0 & 10.3 & 17.4 & 6.7 & 3.8 & 33.3 & 25.0 & 12.0 & 33.3 & 21.4 & 40.0 & 19.2 & 19.0\\

\bottomrule
\end{tabular}

}
\caption{The performance of closed-source models with Image Replacement. Results are measured by Pass@1 (\%).}
\label{tab:prompting_captioning}
\end{table*}

% Gemini Pro & 14.8 & 27.7 & 3.8 & 4.4 & 12.5 & 3.3 & 16.8 & 18.4 & 8.9 & 22.4 & 18.2 & 21.8 & 19.3
% GPT-4 & 32.8 & 51.9 & 18.8 & 30.3 & 20.9 & 17.1 & 41.6 & 38.3 & 38.6 & 45.1 & 25.0 & 42.0 & 36.5

%% file: tables/results_cot.tex
\begin{table*}[t!]
\tiny
\centering
\tabcolsep 2pt
\resizebox{\textwidth}{!}{

\begin{tabular}{lccccccccccccc}
\toprule
\multirow{2}{*}{\textbf{Model}} & \multicolumn{12}{c}{\textbf{Task Type}} & \multirow{2}{*}{\textbf{Average}} \\ 
\cmidrule(lr){2-13}
& \textbf{Linear} & \textbf{Tree} & \textbf{Graph} & \textbf{2D} & \textbf{3D} & \textbf{Chessboard} & \textbf{Map} & \textbf{Math} & \textbf{Patterns} & \textbf{Table} & \textbf{Pseudocode} & \textbf{Others} &  \\ 
\midrule
Gemini Pro Vision & 8.0 & 0.0 & 0.0 & 6.7 & 0.0 & 13.3 & 3.7 & 0.0 & 3.7 & 0.0 & 20.0 & 0.0 & 3.8\\
GPT-4V \scalebox{0.7}{\texttt{(gpt-4-vision-preview)}} & 28.0 & 6.9 & 8.7 & 6.9 & 7.7 & 7.1 & 28.0 & 9.5 & 33.3 & 14.3 & 40.0 & 12.5 & 16.6\\

\bottomrule
\end{tabular}

}
\caption{The performance of closed-source models with Captioning Chain of Thought. Results are measured by Pass@1 (\%).}
\label{tab:prompting_cot}
\end{table*}

% Gemini Pro Vision Test Accuracies & 10.5 & 12.6 & 2.6 & 8.9 & 14.3 & 7.1 & 21.0 & 10.1 & 13.8 & 8.9 & 0.9 & 20.2 & 8.9
% GPT-4V Test Accuracies & 30.0 & 47.1 & 15.2 & 19.8 & 18.4 & 16.9 & 32.3 & 38.3 & 30.9 & 45.1 & 19.6 & 47.0 & 28.0

%% file: figures/error_analysis.tex
\begin{figure*}[t!]

% \begin{adjustwidth}{-1cm}{-1cm}

\centering
    \includegraphics[width=0.9\textwidth]{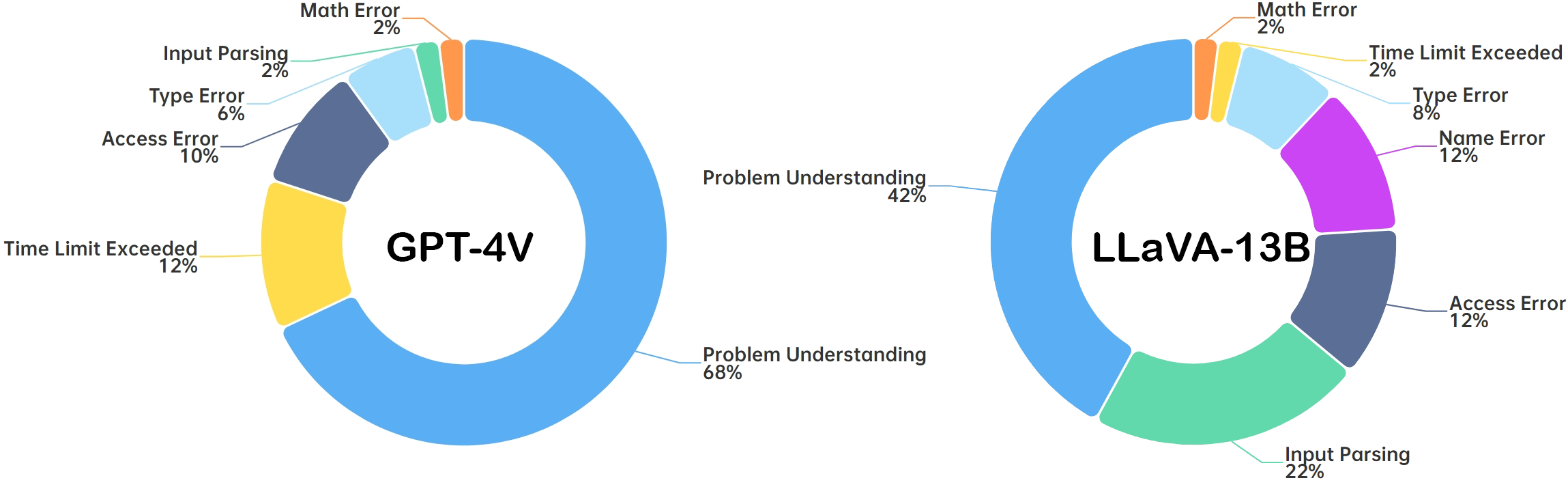}

     % \begin{subfigure}[b]{0.3\textwidth}
     %     \centering
     %     \includegraphics[width=\textwidth]{graph3}
     %     \caption{$y=5/x$}
     %     \label{fig:five over x}
     % \end{subfigure}
        
\caption{Error distribution of GPT-4V and LLaVA-13B on a sampled subset of 50 problems.}
\label{fig:error_analysis}

% \end{adjustwidth}
    
\end{figure*}

%% file: sections/appendix.tex
\appendix
\section{Data Sources}
\label{sec:appendix:data_sources}

The data in MMCode is collected from the following websites\footnote{The license is Apache 2.0 from TACO unless specifically stated.}: 
\begin{itemize}
    \item AtCoder: \url{https://atcoder.jp}
    \item Aizu: \url{https://judge.u-aizu.ac.jp/onlinejudge/}
    \item CodeChef: \url{https://www.codechef.com}
    \item CodeForces: \url{https://codeforces.com}\footnote{No license found.}
    \item CodeWars: \url{https://www.codewars.com}
    \item Project Euler: \url{https://projecteuler.net}\footnote{CC BY-NC-SA 4.0.}
    \item GeeksForGeeks: \url{https://www.geeksforgeeks.org}
    \item HackerRank: \url{https://www.hackerrank.com}
    \item LeetCode: \url{https://leetcode.com}
    \item Open Kattis: \url{https://open.kattis.com/}
\end{itemize}

The statistical data of the quantity of questions and images retained from each platform can be found in Table \ref{tab:data_sources}. In total, MMCode comprises 3548 programming questions. Among the sources, CodeForces and Aizu contribute the most questions and images in MMCode. 
% Table: Data sources
\input{tables/data_sources}

\newpage
\section{Definition of Image Categories}
\label{appendix:sec:category_defs}
\begin{itemize}
\item \textbf{Linear Data Structure:} This category includes diagrams that illustrate sequential data structures such as arrays, linked lists, and queues, where data elements are arranged in a linear order.

\item \textbf{Tree:} Dedicated to the data structure of trees, focusing on hierarchical representations.

\item \textbf{Graph:} Includes visuals of graph data structures where nodes are connected by edges, e.g., directed and undirected graphs. If the problem description is about graphs but the image depicts a tree (e.g. after pruning), it is still classified under this category.

\item \textbf{2D Geometry:} Focuses on two-dimensional geometric shapes and properties, including points, lines, polygons, etc., emphasizing spatial relationships in a plane.

\item \textbf{3D Geometry:} Comprises images that depict three-dimensional objects and structures, such as 3D coordinate systems, orthographic projections, and nets of cubes, showcasing the complexity and characteristics of three-dimensional space.

\item \textbf{Chessboard:} This category includes images showing a chessboard, where the model is expected to solve a problem with respect to some rules of playing.

\item \textbf{Map:} Pertains to images displaying maps that show positions. If the image features a graph functioning as a map, it falls into this category.

\item \textbf{Patterns:} Covers images that involve recognizing, generating, or solving puzzles and patterns, which could be numerical, geometrical, or based on character arrangements.

\item \textbf{Table:} Dedicated to tabular data presentations. 

\item \textbf{Pseudocode:} Includes images that contain pseudocode or simplified code, posing challenges to the dense OCR ability of the models.

\item \textbf{Others:} Serves as a miscellaneous category for visual content that does not fit into other categories, e.g. bar graphs, pie charts, and Venn diagrams.
\end{itemize}

\newpage
\onecolumn
\section{Prompts}
All prompts used in this work are listed in Table \ref{tab:prompts}.
\label{sec:prompts}
\input{tables/prompts}

\input{sections/appendices/data_samples}

\clearpage
\section{More experiments}
\subsection{Test Case Average Pass Rates}
\label{appendix:test_case_avg}

We also report partial success metrics measured by the test case average following APPS[1], presented in Table \ref{appendix:table:test_case_avg}. We observed that it aligns well with the pass@1 reported in Table \ref{tab:main_results}. 
\input{tables/test_case_avg}

\subsection{Image Replacement Captioning}
In this experiment, language-only models are prompted with Image Replacement captioning, but the captions are generated by different models. The results are showcased in Table \ref{tab:caption_combination}. GPT-4's accuracy drops when using Gemini Pro Vision's captions, while Gemini Pro yields identical results.
\input{tables/caption_combination}

\subsection{Image Positions}

Since the problems are typically long, it is uncertain if the images receive sufficient attention of the model. Motivated by the findings of~\citet{liu2024lost-in-the-middle}, we explored whether the position of the images affects the performance.

The results in Table \ref{tab:image_positions} illustrate the impact of image positioning in the problem statements. Specifically, for Gemini Pro Vision, maintaining images in their original positions results in the highest pass rates. Grouping the images at either the beginning or the end of the texts hurt performance. Notably, GPT-4V demonstrates significant robustness, with its overall accuracy remaining unaffected.
\input{tables/image_positions} 

% Captions case study
\input{sections/appendices/caption_case_study}

\input{sections/appendices/code_case_study}

%% file: tables/data_sources.tex
\begin{table}[h!]
\centering
\begin{tabular}{lcc}
\toprule
\textbf{Platform} & \multicolumn{1}{c}{\textbf{\# Questions}} & \multicolumn{1}{c}{\textbf{\# Images}} \\
\midrule
AtCoder       & 139  & 234  \\
Aizu          & 694  & 1349 \\
CodeChef      & 90   & 134 \\
CodeForces    & 1941 & 3837 \\
Project Euler & 132  & 176  \\
GeeksForGeeks & 128  & 192 \\
Open Kattis   & 145  & 195 \\
HackerRank    & 169  & 316  \\
CodeWars      & 33   & 46 \\
LeetCode      & 77   & 141  \\
\midrule
\textbf{Total} & 3548                  & 6620        \\
\bottomrule
\end{tabular}
\caption{The data sources of MMCode and the number of questions and images from each source.}
% with 24730 crawled questions
\label{tab:data_sources}
\end{table}

%% file: tables/prompts.tex
\begin{table*}[h!]
\centering
\begin{tabular}{|m{3cm}|m{10.6cm}|}
\hline
\textbf{Type} & \textbf{Prompt}  \\
\hline
Problem Solving  
& \textbf{System Prompt (if applicable):}\newline 
You are a professional programming contester trying to solve algorithmic problems. The problems come with a description and some images, and you should write a Python solution. \newline 
\newline 
\textbf{User Prompt:}\newline 
You are required to solve a programming problem. Please enclose your code inside a \verb|```python```| block. Do not write a main() function. If a Call-Based format is used, return the result in an appropriate place instead of printing it.\newline 
\newline 
\{\textit{problem statement}\} \newline \\
\hline
Caption Generation 
& Please describe and explain the images in the programming problem. The readers will not be able to see the image, so make sure you include all important information for solving the problem. Please enclose your explanations inside \verb|```plain```| blocks, one for each image. Your output should look like:\newline 
Caption:\newline
\verb|```|plain\newline 
The image shows...\verb|```|\newline 
\newline 
\{\textit{problem statement with only one image}\} \newline\\

\hline
\end{tabular}
\caption{The prompts used in this study.}
\label{tab:prompts}
\end{table*}

%% file: sections/appendices/data_samples.tex
\newpage
\section{Data Samples}

\subsection{An example of a question with a pseudo code image}
\begin{tcolorbox}[breakable] % [everyline=true]

\label{appendix:sec:examples:pseudocode}
% cf_1332_D
Bob is playing a game named "Walk on Matrix".
In this game, player is given an $n \times m$ matrix $A=(a_{i,j})$, i.e. the element in the $i$-th row in the $j$-th column is $a_{i,j}$. Initially, player is located at position $(1,1)$ with score $a_{1,1}$.
To reach the goal, position $(n,m)$, player can move right or down, i.e. move from $(x,y)$ to $(x,y+1)$ or $(x+1,y)$, as long as player is still on the matrix.
However, each move changes player's score to the bitwise AND of the current score and the value at the position he moves to.
Bob can't wait to find out the maximum score he can get using the tool he recently learnt  — dynamic programming. Here is his algorithm for this problem:

\begin{center}
  \includegraphics[width=\textwidth]{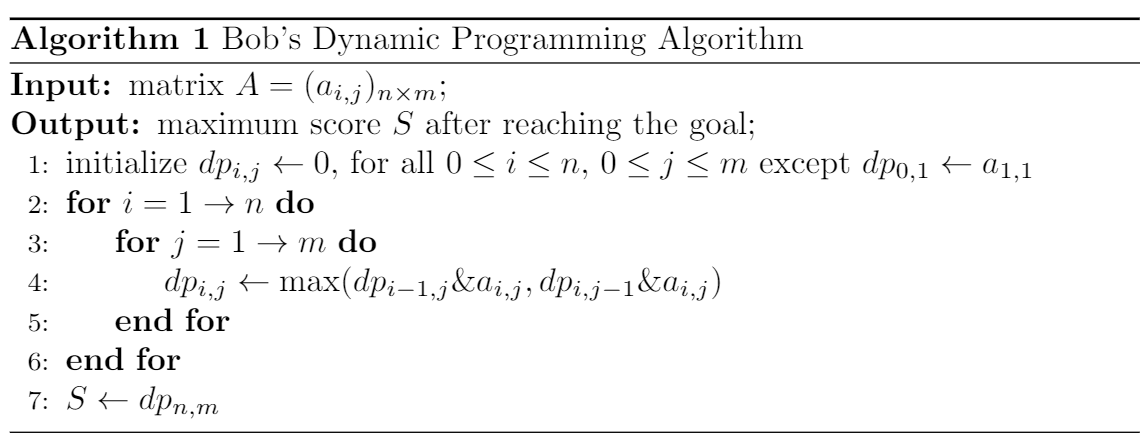}
\end{center}

However, he suddenly realizes that the algorithm above fails to output the maximum score for some matrix $A$. Thus, for any given non-negative integer $k$, he wants to find out an $n \times m$ matrix $A=(a_{i,j})$ such that:

\begin{itemize}
    \item $1 \le n,m \le 500$ (as Bob hates large matrices);
    \item $0 \le a_{i,j} \le 3 \cdot 10^5$ for all $1 \le i\le n,1 \le j\le m$ (as Bob hates large numbers);
    \item the difference between the maximum score he can get and the output of his algorithm is exactly $k$. It can be shown that for any given integer $k$ such that $0 \le k \le 10^5$, there exists a matrix satisfying the above constraints.
\end{itemize}

\textbf{Input}

The only line of the input contains one single integer $k$ ($0 \le k \le 10^5$).

\textbf{Output}

Output two integers $n$, $m$ ($1 \le n,m \le 500$) in the first line, representing the size of the matrix.
Then output $n$ lines with $m$ integers in each line, $a_{i,j}$ in the $(i+1)$-th row, $j$-th column.

\textbf{Examples}

\textbf{Input}
\begin{verbatim}
0
\end{verbatim}

\textbf{Output}
\begin{verbatim}
1 1
300000
\end{verbatim}

\textbf{Input}
\begin{verbatim}
1
\end{verbatim}

\textbf{Output}
\begin{verbatim}
3 4
7 3 3 1
4 8 3 6
7 7 7 3
\end{verbatim}

\textbf{Note}

In the first example, the maximum score Bob can achieve is $300000$, while the output of his algorithm is $300000$.

% \url{https://open.kattis.com/problems/annoyedcoworkers}
\end{tcolorbox}

\subsection{An example of a question with a 2D geometry image}
\begin{tcolorbox}[breakable]

You have most definitely heard the legend of King Arthur and the Knights of the Round Table. Almost all versions of this story proudly point out that the roundness of the Round Table is closely related to Arthur’s belief of equality among the Knights. That is a lie! In fact, Arthur’s choice of table is conditioned by his childhood traumas.
In fact, Arthur was forced to clean up quadratic tables from a young age after a tournament in pick-up sticks had been played on them. After the tournament, typically there would be a bunch of sticks on the table that do not touch each other. In the spirit of the game, the organizers issued strict regulations for the table cleaners. More precisely, the sticks on the table need to be removed one by one in a way that the cleaners pull them in the shortest way towards the edge of the table closest to where they are currently sitting. They also mustn’t rotate or touch the other sticks while doing this (not even in the edge points).
In this task, we will represent the table in the coordinate system with a square that has opposite points in the coordinates $(0,0)$ and $(10\,000, 10\,000)$, whereas the sticks will be represented with straight line segments that lie within that square. We will assume that Arthur is sitting at the edge of the table lying on the $x$-axis. Then the movement of the stick comes down to translating the line segment along the shortest path towards the $x$-axis until the stick falls off the table (as shown in the image). It is your task to help Arthur determine the order of stick movements that meets the requirements from the previous paragraph.
\begin{center}
  \includegraphics[width=0.75\textwidth]{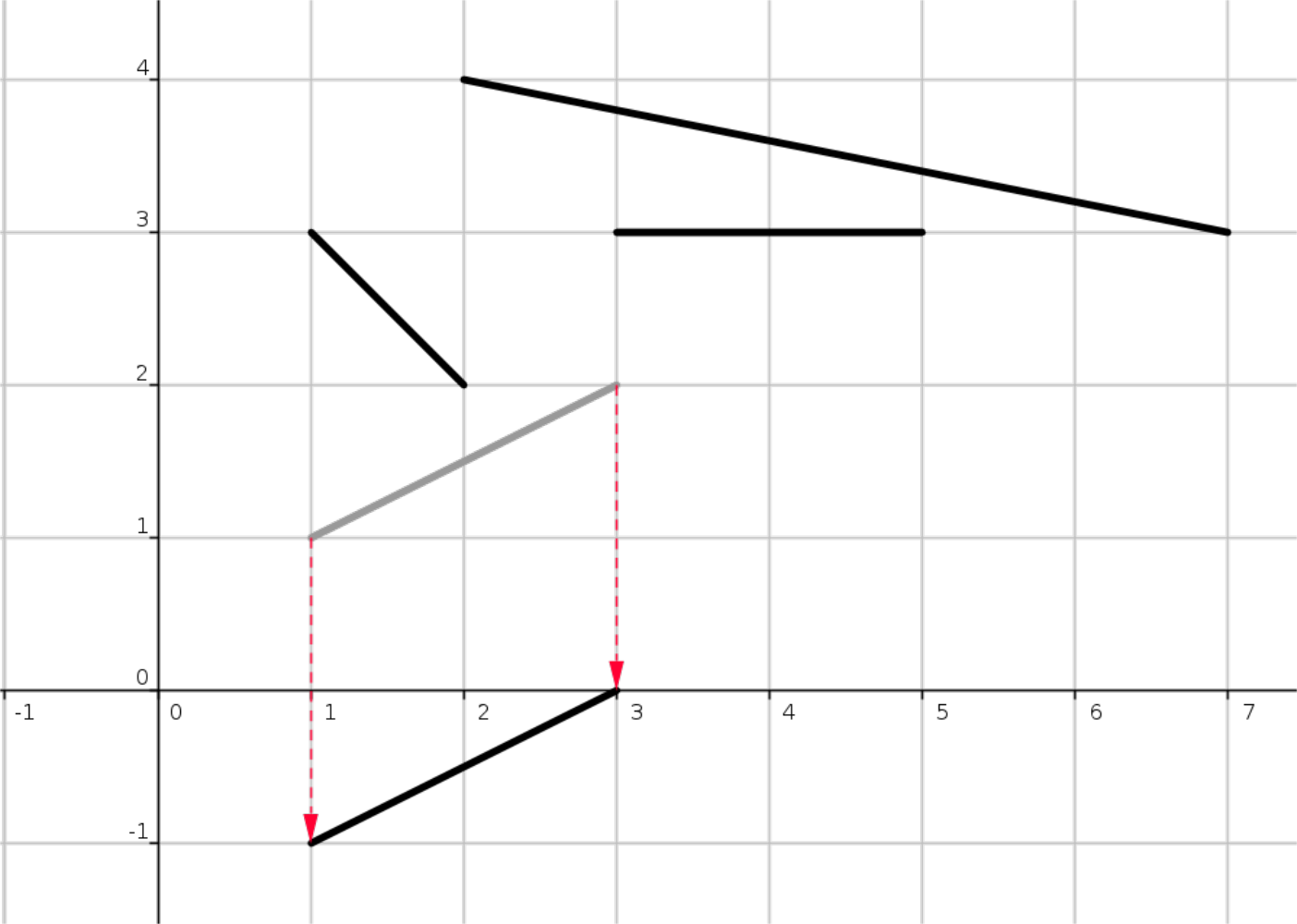}
\end{center}
\textbf{Input}

The first line of input contains the integer $N$ $(1 \leq N \leq 5\, 000)$, the number of sticks on the table. Each of the following $N$ lines contains four integers $x_1$, $y_1$, $x_2$, $y_2$ $(0 \leq x_1, y_1, x_2, y_2 \leq 10\,000)$ that denote the edge points of a stick.

\textbf{Output}

The first and only line of output must contain space-separated stick labels in the order which they need to be taken off the table. A stick’s label corresponds to its position in the input sequence. If there are multiple possible solutions, output any of them.

\textbf{Sample Input 1}
\begin{verbatim}
4
1 3 2 2
1 1 3 2
2 4 7 3
3 3 5 3
\end{verbatim}

\textbf{Sample Output 1}
\begin{verbatim}
2 4 1 3
\end{verbatim}

\textbf{Sample Input 2}
\begin{verbatim}
4
0 0 1 1
1 2 0 3
2 2 3 3
4 0 3 1
\end{verbatim}

\textbf{Sample Output 2}
\begin{verbatim}
4 3 1 2
\end{verbatim}

\textbf{Sample Input 3}
\begin{verbatim}
3
4 6 5 5
2 1 15 1
3 2 8 7
\end{verbatim}

\textbf{Sample Output 3}
\begin{verbatim}
2 3 1
\end{verbatim}

\end{tcolorbox}

\subsection{An example of a question with a 3D geometry image}
\begin{tcolorbox}[breakable]

In AD 3456, the earth is too small for hundreds of billions of people to live in peace. Interstellar Colonization Project with Cubes (ICPC) is a project that tries to move people on the earth to space colonies to ameliorate the problem. ICPC obtained funding from governments and manufactured space colonies very quickly and at low cost using prefabricated cubic blocks.

The largest colony looks like a Rubik's cube. It consists of $3 \times 3 \times 3$ cubic blocks (Figure J.1A). Smaller colonies miss some of the blocks in the largest colony.

When we manufacture a colony with multiple cubic blocks, we begin with a single block. Then we iteratively glue a next block to existing blocks in a way that faces of them match exactly. Every pair of touched faces is glued.
\begin{center}
\includegraphics[width=\textwidth]{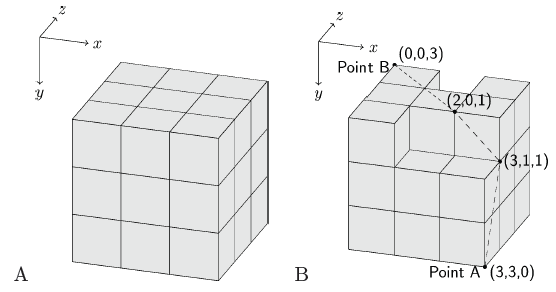}
\\
Figure J.1: Largest colony
\end{center}
However, just before the first launch, we found a design flaw with the colonies. We need to add a cable to connect two points on the surface of each colony, but we cannot change the inside of the prefabricated blocks in a short time. Therefore we decided to attach a cable on the surface of each colony. If a part of the cable is not on the surface, it would be sheared off during the launch, so we have to put the whole cable on the surface. We would like to minimize the lengths of the cables due to budget constraints. The dashed line in Figure J.1B is such an example.

\textbf{Input}

The input contains a series of datasets. Each dataset describes a single colony and the pair of the points for the colony in the following format.

$x_{1} y_{1} z_{1} x_{2} y_{2} z_{2}$\\
$b_{0,0,0}b_{1,0,0}b_{2,0,0}$\\
$b_{0,1,0}b_{1,1,0}b_{2,1,0}$\\
$b_{0,2,0}b_{1,2,0}b_{2,2,0}$\\
$b_{0,0,1}b_{1,0,1}b_{2,0,1}$\\
$b_{0,1,1}b_{1,1,1}b_{2,1,1}$\\
$b_{0,2,1}b_{1,2,1}b_{2,2,1}$\\
$b_{0,0,2}b_{1,0,2}b_{2,0,2}$\\
$b_{0,1,2}b_{1,1,2}b_{2,1,2}$\\
$b_{0,2,2}b_{1,2,2}b_{2,2,2}$\\

$(x_{1}, y_{1}, z_{1})$ and $(x_{2}, y_{2}, z_{2})$ are the two distinct points on the surface of the colony, where $x_{1}, x_{2}, y_{1}, y_{2}, z_{1}, z_{2}$ are integers that satisfy $0 \leq x_{1}, x_{2}, y_{1}, y_{2}, z_{1}, z_{2} \leq 3$. $b_{i,j,k}$ is '\#' when there is a cubic block whose two diagonal vertices are $(i, j, k)$ and $(i + 1, j + 1, k + 1)$, and $b_{i,j,k}$ is '.' if there is no block. Figure J.1A corresponds to the first dataset in the sample input, whereas Figure J.1B corresponds to the second. A cable can pass through a zero-width gap between two blocks if they are touching only on their vertices or edges. In Figure J.2A, which is the third dataset in the sample input, the shortest cable goes from the point A (0, 0, 2) to the point B (2, 2, 2), passing through (1, 1, 2), which is shared by six blocks. Similarly, in Figure J.2B (the fourth dataset in the sample input), the shortest cable goes through the gap between two blocks not glued directly. When two blocks share only a single vertex, you can put a cable through the vertex (Figure J.2C; the fifth dataset in the sample input).

You can assume that there is no colony consisting of all $3 \times 3 \times 3$ cubes but the center cube.

Six zeros terminate the input.
\begin{center}
\includegraphics[width=\textwidth]{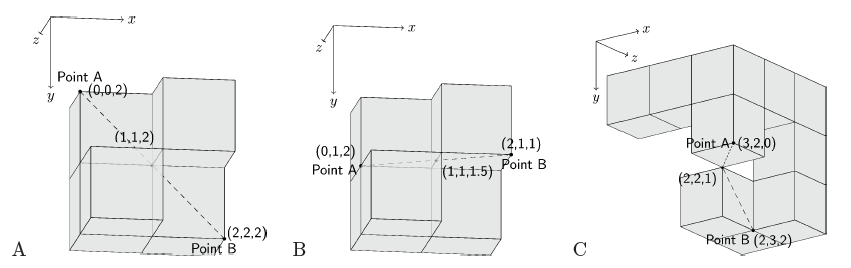}
\\
Figure J.2: Largest colony
\end{center}
\textbf{Output}

For each dataset, output a line containing the length of the shortest cable that connects the two given points. We accept errors less than $0.0001$. You can assume that given two points can be connected by a cable.

\textbf{Sample Input}

\begin{verbatim}
0 0 0 3 3 3
###
###
###
###
###
###
###
###
###
3 3 0 0 0 3
#..
###
###
###
###
###
#.#
###
###
0 0 2 2 2 2
...
...
...
.#.
#..
...
##.
##.
...
0 1 2 2 1 1
...
...
...
.#.
#..
...
##.
##.
...
3 2 0 2 3 2
###
..#
...
..#
...
.#.
..#
..#
.##
0 0 0 0 0 0
\end{verbatim}

\textbf{Output for the Sample Input}

\begin{verbatim}
6.70820393249936941515
6.47870866461907457534
2.82842712474619029095
2.23606797749978980505
2.82842712474619029095
\end{verbatim}

\end{tcolorbox}

\subsection{An example of a question with a tree image}
\begin{tcolorbox}[breakable]

Let's define the Eulerian traversal of a tree (a connected undirected graph without cycles) as follows: consider a depth-first search algorithm which traverses vertices of the tree and enumerates them in the order of visiting (only the first visit of each vertex counts). This function starts from the vertex number $1$ and then recursively runs from all vertices which are connected with an edge with the current vertex and are not yet visited in increasing numbers order. Formally, you can describe this function using the following pseudocode:

\begin{verbatim}
next_id = 1
id = array of length n filled with -1
visited = array of length n filled with false

function dfs(v):
    visited[v] = true
    id[v] = next_id
    next_id += 1
    for to in neighbors of v in increasing order:
        if not visited[to]:
            dfs(to)
\end{verbatim}

You are given a weighted tree, the vertices of which were enumerated with integers from $1$ to $n$ using the algorithm described above.

A leaf is a vertex of the tree which is connected with only one other vertex. In the tree given to you, the vertex $1$ is not a leaf. The distance between two vertices in the tree is the sum of weights of the edges on the simple path between them.

You have to answer $q$ queries of the following type: given integers $v$, $l$ and $r$, find the shortest distance from vertex $v$ to one of the leaves with indices from $l$ to $r$ inclusive.

\textbf{Input}

The first line contains two integers $n$ and $q$ ($3 \leq n \leq 500\,000$, $1 \leq q \leq 500\,000$) — the number of vertices in the tree and the number of queries, respectively.

The $(i - 1)$-th of the following $n - 1$ lines contains two integers $p_i$ and $w_i$ ($1 \leq p_i < i$, $1 \leq w_i \leq 10^9$), denoting an edge between vertices $p_i$ and $i$ with the weight $w_i$.

It's guaranteed that the given edges form a tree and the vertices are enumerated in the Eulerian traversal order and that the vertex with index $1$ is not a leaf.

The next $q$ lines describe the queries. Each of them contains three integers $v_i$, $l_i$, $r_i$ ($1 \leq v_i \leq n$, $1 \leq l_i \leq r_i \leq n$), describing the parameters of the query. It is guaranteed that there is at least one leaf with index $x$ such that $l_i \leq x \leq r_i$.

\textbf{Output}

Output $q$ integers — the answers for the queries in the order they are given in the input.

\textbf{Examples}

\textbf{Input}

\begin{verbatim}
5 3
1 10
1 1
3 2
3 3
1 1 5
5 4 5
4 1 2
\end{verbatim}

\textbf{Output}

\begin{verbatim}
3
0
13
\end{verbatim}

\textbf{Input}

\begin{verbatim}
5 3
1 1000000000
2 1000000000
1 1000000000
1 1000000000
3 4 5
2 1 5
2 4 5
\end{verbatim}

\textbf{Output}

\begin{verbatim}
3000000000
1000000000
2000000000
\end{verbatim}

\textbf{Input}

\begin{verbatim}
11 8
1 7
2 1
1 20
1 2
5 6
6 2
6 3
5 1
9 10
9 11
5 1 11
1 1 4
9 4 8
6 1 4
9 7 11
9 10 11
8 1 11
11 4 5
\end{verbatim}

\textbf{Output}

\begin{verbatim}
8
8
9
16
9
10
0
34
\end{verbatim}

\textbf{Note}

In the first example, the tree looks like this:

\begin{center}
\includegraphics[width=0.3\textwidth]{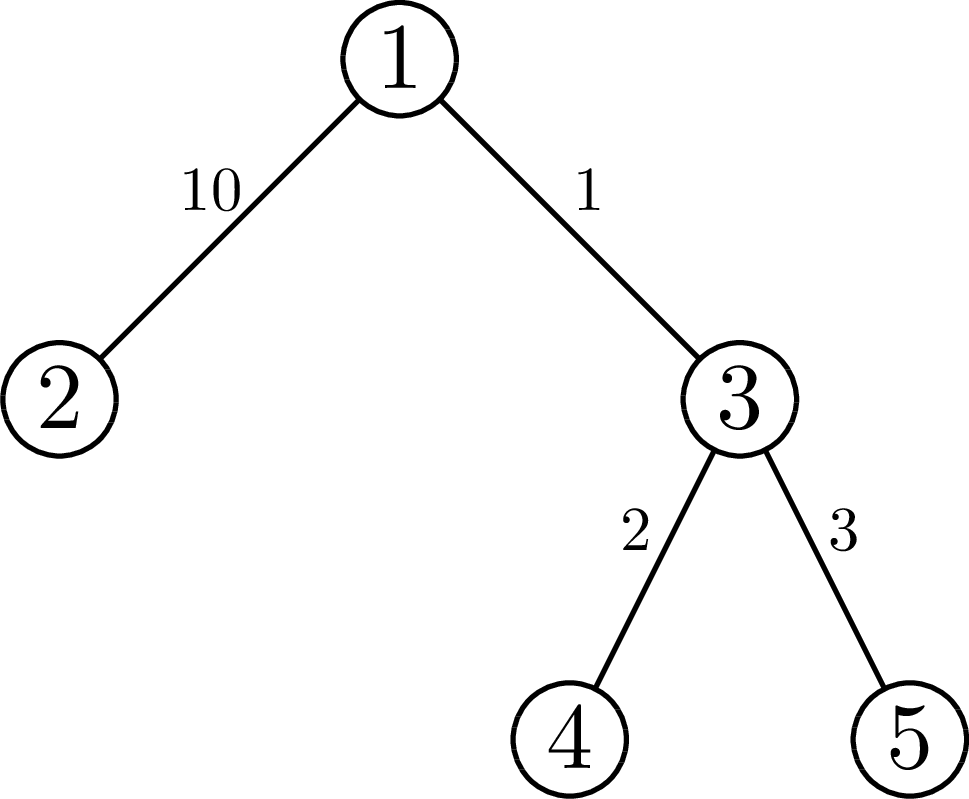}
\end{center}

In the first query, the nearest leaf for the vertex $1$ is vertex $4$ with distance $3$. In the second query, the nearest leaf for vertex $5$ is vertex $5$ with distance $0$. In the third query, the nearest leaf for vertex $4$ is vertex $4$; however, it is not inside interval $[1, 2]$ of the query. The only leaf in interval $[1, 2]$ is vertex $2$ with distance $13$ from vertex $4$.

\end{tcolorbox}

\subsection{An example of a question with a graph image}
\begin{tcolorbox}[breakable]
Fox Ciel just designed a puzzle game called "Polygon"! It is played using triangulations of a regular $n$-edge polygon. The goal is to transform one triangulation to another by some tricky rules.
\begin{center}
\includegraphics[width=\textwidth]{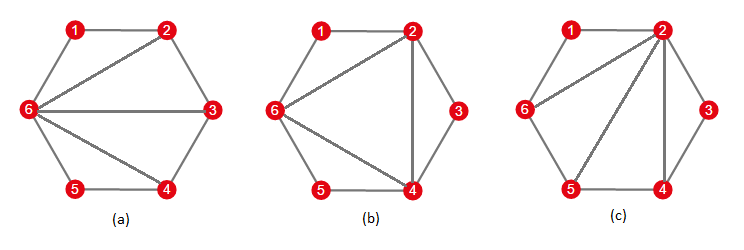}
\end{center}
Triangulation of an $n$-edge polygon is a set of $n - 3$ diagonals satisfying the condition that no two diagonals share a common internal point.

For example, the initial state of the game may look like (a) in the figure. And your goal may look like (c). In each step, you can choose a diagonal inside the polygon (but not one of the edges of the polygon) and flip this diagonal.

Suppose you are going to flip a diagonal $a - b$. There always exist two triangles sharing $a - b$ as a side, let's denote them as $a - b - c$ and $a - b - d$. As a result of this operation, the diagonal $a - b$ is replaced by a diagonal $c - d$. It can be easily proven that after the flip operation, the resulting set of diagonals is still a triangulation of the polygon.

So in order to solve the above case, you may first flip diagonal $6 - 3$, it will be replaced by diagonal $2 - 4$. Then you flip diagonal $6 - 4$ and get figure (c) as a result.

Ciel just proved that for any starting and destination triangulations, this game has a solution. She wants you to solve it in no more than $20,000$ steps for any puzzle satisfying $n \leq 1000$.

\textbf{Input}

The first line contains an integer $n$ ($4 \leq n \leq 1000$), the number of edges of the regular polygon.

Then follows two groups of $(n - 3)$ lines describing the original triangulation and goal triangulation.

Description of each triangulation consists of $(n - 3)$ lines. Each line contains $2$ integers $a_{i}$ and $b_{i}$ ($1 \leq a_{i}, b_{i} \leq n$), describing a diagonal $a_{i} - b_{i}$.

It is guaranteed that both original and goal triangulations are correct (i.e., no two diagonals share a common internal point in both of these triangulations).

\textbf{Output}

First, output an integer $k$ ($0 \leq k \leq 20,000$): the number of steps.

Then output $k$ lines, each containing $2$ integers $a_{i}$ and $b_{i}$: the endpoints of a diagonal you are going to flip at step $i$. You may output $a_{i}$ and $b_{i}$ in any order.

If there are several possible solutions, output any of them.

\textbf{Examples}

\textbf{Input}

\begin{verbatim}
41 32 4
\end{verbatim}

\textbf{Output}

\begin{verbatim}
11 3
\end{verbatim}

\textbf{Input}

\begin{verbatim}
62 63 64 66 25 24 2
\end{verbatim}

\textbf{Output}

\begin{verbatim}
26 36 4
\end{verbatim}

\textbf{Input}

\begin{verbatim}
87 12 77 36 34 66 16 26 36 46 8
\end{verbatim}

\textbf{Output}

\begin{verbatim}
37 37 27 1
\end{verbatim}

\textbf{Note}

Sample test $2$ is discussed above and shown on the picture.

\end{tcolorbox}

\subsection{An example of a question with an unrelated image}
\begin{tcolorbox}[breakable]

Bash got tired on his journey to become the greatest Pokemon master. So he decides to take a break and play with functions.

Bash defines a function $f_{0}(n)$, which denotes the number of ways of factoring $n$ into two factors $p$ and $q$ such that $\text{gcd}(p, q) = 1$. In other words, $f_{0}(n)$ is the number of ordered pairs of positive integers $(p, q)$ such that $p \cdot q = n$ and $\text{gcd}(p, q) = 1$.

But Bash felt that it was too easy to calculate this function. So he defined a series of functions, where $f_{r+1}$ is defined as:

\begin{center}
\includegraphics[width=0.3\textwidth]{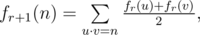}
\end{center}

Where $(u, v)$ is any ordered pair of positive integers, they need not to be co-prime.

Now Bash wants to know the value of $f_{r}(n)$ for different $r$ and $n$. Since the value could be huge, he would like to know the value modulo $10^9 + 7$. Help him!

\textbf{Input}

The first line contains an integer $q$ ($1 \leq q \leq 10^6$) — the number of values Bash wants to know.

Each of the next $q$ lines contains two integers $r$ and $n$ ($0 \leq r \leq 10^6$, $1 \leq n \leq 10^6$), which denote Bash wants to know the value $f_{r}(n)$.

\textbf{Output}

Print $q$ integers. For each pair of $r$ and $n$ given, print $f_{r}(n)$ modulo $10^9 + 7$ on a separate line.

\textbf{Example}

\textbf{Input}

\begin{verbatim}
50
301 253
652 54
48
\end{verbatim}

\textbf{Output}

\begin{verbatim}
85254630
\end{verbatim}
\end{tcolorbox}

\subsection{An example of a question with a table image}
\begin{tcolorbox}[breakable]

At a regular competition, Vladik and Valera won $a$ and $b$ candies respectively. Vladik offered 1 his candy to Valera. After that, Valera gave Vladik 2 his candies, so that no one thought that he was less generous. Vladik for the same reason gave 3 candies to Valera in the next turn.

More formally, the guys take turns giving each other one candy more than they received in the previous turn.

This continued until the moment when one of them couldn’t give the right amount of candy. Candies, which guys got from each other, they don’t consider as their own. You need to know who is the first who can’t give the right amount of candy.

\textbf{Input}

A single line of input data contains two space-separated integers $a, b$ $(1 \leq a, b \leq 10^9)$ — the number of Vladik and Valera candies respectively.

\textbf{Output}

Print a single line "Vladik" if Vladik is the first who can’t give the right amount of candy, or "Valera" otherwise.

\textbf{Examples}

\textbf{Input}

\begin{verbatim}
1 1
\end{verbatim}

\textbf{Output}

\begin{verbatim}
Valera
\end{verbatim}

\textbf{Input}

\begin{verbatim}
7 6
\end{verbatim}

\textbf{Output}

\begin{verbatim}
Vladik
\end{verbatim}

\textbf{Note}

Illustration for the first test case:
\begin{center}
\includegraphics[width=0.3\textwidth]{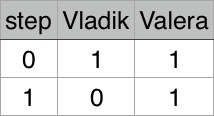}
\end{center}
Illustration for the second test case:
\begin{center}
\includegraphics[width=0.3\textwidth]{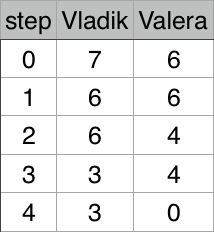}
\end{center}
\end{tcolorbox}

\subsection{An example of a question with an linear data structure image}
\begin{tcolorbox}[breakable]

Dima is a beginner programmer. During his working process, he regularly has to repeat the following operation again and again: to remove every second element from the array. One day he has been bored with easy solutions of this problem, and he has come up with the following extravagant algorithm.

Let's consider that initially, the array contains $n$ numbers from $1$ to $n$ and the number $i$ is located in the cell with the index $2i - 1$ (Indices are numbered starting from one) and other cells of the array are empty. Each step Dima selects a non-empty array cell with the maximum index and moves the number written in it to the nearest empty cell to the left of the selected one. The process continues until all $n$ numbers will appear in the first $n$ cells of the array. For example if $n = 4$, the array is changing as follows:

\begin{center}
\includegraphics[width=\textwidth]{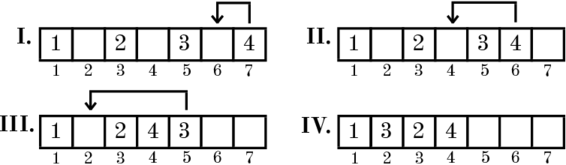}
\end{center}

You have to write a program that allows you to determine what number will be in the cell with index $x$ $(1 \leq x \leq n)$ after Dima's algorithm finishes.

\textbf{Input}

The first line contains two integers $n$ and $q$ $(1 \leq n \leq 10^{18}, 1 \leq q \leq 200,000)$, the number of elements in the array and the number of queries for which it is needed to find the answer.

Next $q$ lines contain integers $x_i$ $(1 \leq x_i \leq n)$, the indices of cells for which it is necessary to output their content after Dima's algorithm finishes.

\textbf{Output}

For each of $q$ queries, output one integer number, the value that will appear in the corresponding array cell after Dima's algorithm finishes.

\textbf{Examples}

\textbf{Input}

\begin{verbatim}
4 3234
\end{verbatim}

\textbf{Output}

\begin{verbatim}
324
\end{verbatim}

\textbf{Input}

\begin{verbatim}
13 410548
\end{verbatim}

\textbf{Output}

\begin{verbatim}
13389
\end{verbatim}

\textbf{Note}

The first example is shown in the picture.

In the second example, the final array is $[1, 12, 2, 8, 3, 11, 4, 9, 5, 13, 6, 10, 7]$.
\end{tcolorbox}

\subsection{An example of a question with an other image}
\begin{tcolorbox}[breakable]

Even polar bears feel cold when lying on the ice. Therefore, a polar bear Alice is going to make a carpet. The carpet can be viewed as a grid with height $h$ and width $w$. Then the grid is divided into $h \times w$ squares. Alice is going to assign one of $k$ different colors to each square. The colors are numbered from $1$ to $k$. She may choose not to use all of the colors.

However, there are some restrictions. For every two adjacent squares (squares that share an edge) $x$ and $y$, there is a color constraint in one of the forms:
\begin{itemize}
    \item color($x$) $=$ color($y$), or
    \item color($x$) $\neq$ color($y$).
\end{itemize}
Example of the color constraints:

\begin{center}
\includegraphics[width=\textwidth]{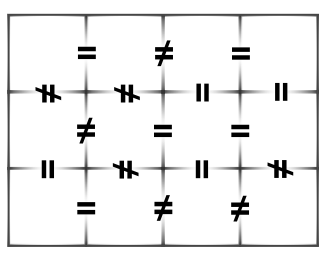}
\end{center}

Ideally, Alice wants to satisfy all color constraints. But again, life in the Arctic is hard. It is not always possible to satisfy all color constraints. Fortunately, she will still be happy if at least $\frac{3}{4}$ of the color constraints are satisfied.

If she has $4$ colors she can color the carpet in the following way:

\begin{center}
\includegraphics[width=\textwidth]{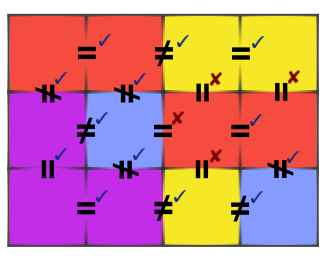}
\end{center}

And she is happy because $\frac{13}{17}$ of the color constraints are satisfied, and $\frac{13}{17} > \frac{3}{4}$. Your task is to help her color the carpet.

\textbf{Input}

The first line contains three integers $h, w, k$ $(2 \leq h, w \leq 1000, 1 \leq k \leq w \cdot h)$.
The next $2h - 1$ lines describe the color constraints from top to bottom, left to right. They contain $w - 1, w, w - 1, w, \ldots, w - 1$ characters respectively. Each color constraint is represented by a character "E" or "N", where "E" means " $=$ " and "N" means " $\neq$ ".
The color constraints are listed in the order they are depicted in the picture.

\textbf{Output}

If there is a coloring that satisfies at least $\frac{3}{4}$ of the color constraints, print "YES" (without quotes) on the first line. In each of the next $h$ lines, print $w$ integers describing the coloring.
Otherwise, print "NO" (without quotes).

\textbf{Examples}

\textbf{Input}

\begin{verbatim}
3 4 4ENENNEENEEENENENN
\end{verbatim}

\textbf{Output}

\begin{verbatim}
YES
1 1 2 2
3 4 1 1
3 3 2 4
\end{verbatim}

\end{tcolorbox}

\subsection{An example of a question with an unrelated image}
\begin{tcolorbox}[breakable]

\label{appendix:sec:unrelated_img}
\begin{wrapfigure}{r}{0.4\textwidth} % "r" for right, "l" for left
  \centering
  \includegraphics[width=0.4\textwidth]{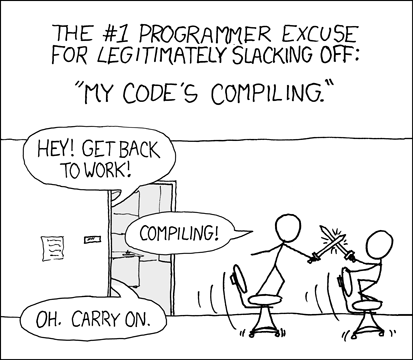}
  \label{fig:unrelated_image}
\end{wrapfigure}

It’s another day in the office, and you’re a mastermind of not doing any work yourself. Instead, you’ll go to your coworkers for “help,” but secretly have them do all the work.

You’ve determined that the more one of your coworkers helps you, the more annoyed they become. You’ve also been able to determine how much more annoyed a coworker gets every time you ask them for help. At the beginning of the day, a coworker is initially $a$ annoyed at you. That’s their annoyance level. Every time you ask them for help though, they become $d$ more annoyed at you – their annoyance level $a$ increases by a constant amount $d$ so that $a = a + d$.

You want to complete a project of $h$ tasks solely with “help” from your coworkers, but you need to be careful not to annoy any of them too much.

What’s the best you can do?

% \url{https://open.kattis.com/problems/annoyedcoworkers}
\end{tcolorbox}

%% file: tables/test_case_avg.tex
\begin{table*}[t!]
\tiny
\centering
\tabcolsep 2pt
\resizebox{\textwidth}{!}{

\begin{tabular}{lccccccccccccc}
\toprule
\textbf{Model} &
\textbf{Linear} &
\textbf{Tree} &
\textbf{Graph} &
\textbf{2D} &
\textbf{3D} &
\textbf{\begin{tabular}[c]{@{}c@{}}Chess-\\ board\end{tabular}} &
\textbf{Map} &
\textbf{Math} &
\textbf{Patterns} &
\textbf{Table} &
\textbf{\begin{tabular}[c]{@{}c@{}}Pseudo-\\ code\end{tabular}} &
\textbf{Others} &
\textbf{Average} \\
\midrule
\multicolumn{14}{c}{\textbf{Language Only Inputs}} \\
LLaVA-1.5-7B (text-only) & 8.1 & 2.1 & 1.3 & 0.6 & 0.1 & 6.7 & 0.0 & 0.2 & 2.9 & 1.8 & 0.0 & 2.7 & 2.2 \\
LLaVA-1.5-13B (text-only) & 9.8 & 0.4 & 3.9 & 0.4 & 0.6 & 0.1 & 0.1 & 4.2 & 4.0 & 1.8 & 11.1 & 1.7 & 2.9 \\
QWEN-VL (text-only) & 4.3 & 1.9 & 0.2 & 2.5 & 3.0 & 9.3 & 3.8 & 1.7 & 0.0 & 1.8 & 10.2 & 2.2 & 2.7 \\
CodeGemma-7b-Instruct & 20.5 & 7.1 & 4.8 & 6.6 & 10.3 & 14.3 & 10.8 & 7.5 & 18.1 & 8.3 & 20.0 & 7.5 & 11.1 \\
CodeLLaMA-7b-instsruct & 12.1 & 3.9 & 5.4 & 5.6 & 0.4 & 3.6 & 1.7 & 5.9 & 8.7 & 6.5 & 10.3 & 0.9 & 5.4 \\
CodeLLaMA-13b-instruct & 9.3 & 7.9 & 5.0 & 5.3 & 2.8 & 6.8 & 2.1 & 1.7 & 3.8 & 5.4 & 10.0 & 1.4 & 4.9 \\
DeepSeekCoder-7b-instruct & 27.4 & 10.8 & 13.8 & 6.4 & 9.8 & 38.0 & 20.9 & 17.4 & 15.7 & 12.7 & 12.4 & 12.7 & 16.5 \\
DeepSeekCoder-33b-instruct & 31.1 & 7.2 & 19.7 & 22.8 & 13.5 & 31.0 & 26.6 & 18.7 & 18.5 & 15.5 & 31.1 & 25.3 & 21.5 \\
LLaMA3-instruct & 17.8 & 4.2 & 10.9 & 10.4 & 7.3 & 11.1 & 11.6 & 7.7 & 7.9 & 22.8 & 1.2 & 6.0 & 9.9 \\
MagiCoder & 24.9 & 8.4 & 17.1 & 8.6 & 3.0 & 20.4 & 17.5 & 10.4 & 21.5 & 12.8 & 21.7 & 19.5 & 15.3 \\
StarCoder-15b-instruct & 16.3 & 5.0 & 4.9 & 12.6 & 4.4 & 2.5 & 3.6 & 7.2 & 10.3 & 5.7 & 10.7 & 2.9 & 7.5 \\
WizardCoder & 11.2 & 5.4 & 6.1 & 8.2 & 2.2 & 10.7 & 6.3 & 10.1 & 15.3 & 5.9 & 21.0 & 6.2 & 8.6 \\
\hdashline
Gemini Pro & 25.0 & 3.9 & 4.9 & 7.5 & 6.0 & 13.6 & 13.2 & 4.3 & 21.1 & 8.1 & 20.4 & 17.1 & 12.0 \\
GPT-3.5-turbo-1106 & 38.3 & 18.6 & 17.9 & 17.1 & 14.6 & 25.4 & 20.8 & 16.8 & 26.9 & 19.0 & 21.9 & 18.6 & 21.0 \\
GPT-4 & 40.3 & 22.0 & 20.9 & 19.5 & 17.7 & 29.6 & 25.3 & 24.7 & 48.4 & 24.1 & 33.8 & 37.2 & 28.0 \\
GPT-4V (text-only) & 52.4 & 17.9 & 23.1 & 20.0 & 17.4 & 37.9 & 23.5 & 20.3 & 34.4 & 30.4 & 44.9 & 39.6 & 28.5 \\
GPT-4o (text-only) & 40.8 & 10.1 & 15.8 & 18.3 & 14.6 & 31.5 & 16.2 & 28.7 & 30.8 & 8.8 & 45.1 & 29.2 & 23.3 \\

\midrule
\multicolumn{14}{c}{\textbf{Vision + Language Inputs}} \\
LLaVA-1.5-7B & 12.6 & 4.5 & 0.6 & 3.4 & 2.7 & 6.7 & 0.8 & 0.1 & 0.2 & 2.6 & 0.0 & 0.4 & 3.2 \\
LLaVA-1.5-13B & 8.2 & 3.9 & 0.1 & 0.1 & 0.6 & 0.7 & 3.2 & 1.0 & 3.7 & 1.8 & 0.7 & 0.3 & 2.3 \\
QWEN-VL & 11.1 & 2.3 & 0.2 & 0.7 & 0.0 & 2.8 & 0.1 & 2.3 & 4.9 & 3.7 & 0.0 & 1.6 & 2.5 \\

\hdashline
Gemini Pro Vision & 20.6 & 4.9 & 5.9 & 7.1 & 6.9 & 10.8 & 15.0 & 5.0 & 16.2 & 7.9 & 31.2 & 7.5 & 10.7 \\
GPT-4V & 59.7 & 22.9 & 21.3 & 19.1 & 19.8 & 37.2 & 26.5 & 16.2 & 39.2 & 24.8 & 43.3 & 29.8 & 29.5 \\
GPT-4o & 44.7 & 12.8 & 20.0 & 13.0 & 15.9 & 34.9 & 34.2 & 31.6 & 37.3 & 15.2 & 50.0 & 25.5 & 27.0 \\

\bottomrule
\end{tabular}

}
\caption{Test case average grouped by different image categories.}
\label{appendix:table:test_case_avg}
\end{table*}

%% file: tables/caption_combination.tex
\begin{table*}[h!]
\centering
\tabcolsep 2pt
\resizebox{\textwidth}{!}{

\begin{tabular}{llccccccccccccc}
\toprule
\textbf{Model} &
\textbf{Caption Model} &
\textbf{Linear} &
\textbf{Tree} &
\textbf{Graph} &
\textbf{2D} &
\textbf{3D} &
\textbf{\begin{tabular}[c]{@{}c@{}}Chess-\\ board\end{tabular}} &
\textbf{Map} &
\textbf{Math} &
\textbf{Patterns} &
\textbf{Table} &
\textbf{\begin{tabular}[c]{@{}c@{}}Pseudo-\\ code\end{tabular}} &
\textbf{Others} &
\textbf{Average} \\
\midrule
Gemini Pro & Gemini Pro Vision & 16.0 & 0.0 & 0.0 & 6.7 & 0.0 & 6.7 & 3.6 & 0.0 & 11.1 & 7.1 & 20.0 & 7.7 & 6.1 \\
Gemini Pro & GPT-4V \tiny{(gpt-4-1106-preview)} & 16.0 & 0.0 & 0.0 & 6.7 & 0.0 & 6.7 & 3.6 & 0.0 & 11.1 & 7.1 & 20.0 & 7.7 & 6.1 \\
GPT-4 \tiny{(gpt-4-1106-preview)} & Gemini Pro Vision & 32.0 & 3.4 & 17.4 & 6.7 & 7.7 & 33.3 & 25.0 & 12.0 & 25.9 & 21.4 & 40.0 & 19.2 & 18.6 \\
GPT-4 \tiny{(gpt-4-1106-preview)} & GPT-4V \tiny{(gpt-4-1106-preview)} & 32.0 & 10.3 & 17.4 & 6.7 & 3.8 & 33.3 & 25.0 & 12.0 & 33.3 & 21.4 & 40.0 & 19.2 & 19.0 \\
\bottomrule
\end{tabular}
}
\caption{Image Replacement captioning performance measured by Pass@1 (\%) of models with different caption sources.}
\label{tab:caption_combination}
\end{table*}

%% file: tables/image_positions.tex
\begin{table*}[h!]
\centering
\tabcolsep 2pt
\resizebox{\textwidth}{!}{

\begin{tabular}{lcccccccccccccc}
\toprule
\textbf{Model} &
\textbf{Image position} &
\textbf{Linear} &
\textbf{Tree} &
\textbf{Graph} &
\textbf{2D} &
\textbf{3D} &
\textbf{\begin{tabular}[c]{@{}c@{}}Chess-\\ board\end{tabular}} &
\textbf{Map} &
\textbf{Math} &
\textbf{Patterns} &
\textbf{Table} &
\textbf{\begin{tabular}[c]{@{}c@{}}Pseudo-\\ code\end{tabular}} &
\textbf{Others} &
\textbf{Total} \\
\midrule
\multirow{3}{*}{Gemini Pro Vision} & in-place & 12.5 & 0.0 & 4.3 & 0.0 & 3.8 & 6.7 & 7.1 & 0.0 & 7.4 & 0.0 & 30.0 & 0.0 & 5.0 \\
& front & 8.7 & 0.0 & 0.0 & 0.0 & 0.0 & 0.0 & 7.1 & 0.0 & 3.7 & 0.0 & 30.0 & 3.8 & 3.4 \\
& end & 16.0 & 0.0 & 0.0 & 0.0 & 3.8 & 0.0 & 3.6 & 0.0 & 7.4 & 0.0 & 30.0 & 3.8 & 4.6 \\
\midrule
\multirow{3}{*}{GPT-4V \tiny{(gpt-4-1106-preview)}} & in-place & 40.0 & 6.9 & 13.0 & 13.8 & 3.8 & 21.4 & 24.0 & 9.5 & 25.9 & 21.4 & 40.0 & 20.8 & 19.4 \\
& front & 36.0 & 6.9 & 8.7 & 6.9 & 7.7 & 50.0 & 24.0 & 0.0 & 33.3 & 14.3 & 40.0 & 20.8 & 19.4 \\
& end & 40.0 & 6.9 & 13.0 & 10.3 & 7.7 & 21.4 & 36.0 & 14.3 & 18.5 & 14.3 & 40.0 & 20.8 & 19.4 \\
\bottomrule
\end{tabular}
}
\caption{The impact of image positioning. ``in-place'' refers to images interspersed within the text, while ``front'' and ``end'' indicate images grouped at the start or end of the problem statement.}
\label{tab:image_positions}
\end{table*}

%% file: sections/appendices/caption_case_study.tex
\clearpage
\section{Case Study: Captions}
\label{appendix:sec:case_study_captions}

\begin{figure}[h!]
    \centering
\begin{tcolorbox}[title=Linear Data Structure]
    \includegraphics[width=\textwidth]{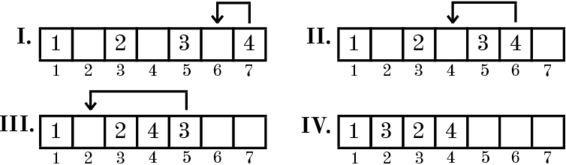}

    \begin{tcolorbox}[breakable, title=Gemini Vision Pro]
    The image shows an array of 4 elements, with indices ranging from 1 to 7. \red{The elements are arranged as follows:\\ \lbrack 1, 2, 3, 4, \_, \_, \_\rbrack \\}
        
    \end{tcolorbox}
    
    \begin{tcolorbox}[breakable, title=GPT-4V]
    The image shows a sequence of four diagrams labeled I. to IV., representing the steps of Dima's algorithm applied to an array with n = 4 elements. Each diagram depicts an array with seven cells, numbered from 1 to 7 at the bottom. Initially, the numbers 1 to 4 are placed in cells 1, 3, 5, and 7, respectively. The diagrams illustrate the process of moving the numbers to the left into the nearest empty cells until all numbers are in the first four cells of the array. Arrows indicate the movement of numbers from one cell to another.
        
    \end{tcolorbox}
\end{tcolorbox}

    \caption{An example of a \texttt{Linear Data Structure} image. Gemini Pro Vision only sees one subfigure, and generates unusable captions. GPT-4V's caption is correct, but the details of step \uppercase\expandafter{\romannumeral 2}, \uppercase\expandafter{\romannumeral 3} and \uppercase\expandafter{\romannumeral 4} are not included.}
    \label{fig:appendix:caption:linear}
\end{figure}

\begin{figure}[h!]
    \centering
\begin{tcolorbox}[title=Pseudocode]
\centering
    \includegraphics[width=\textwidth]{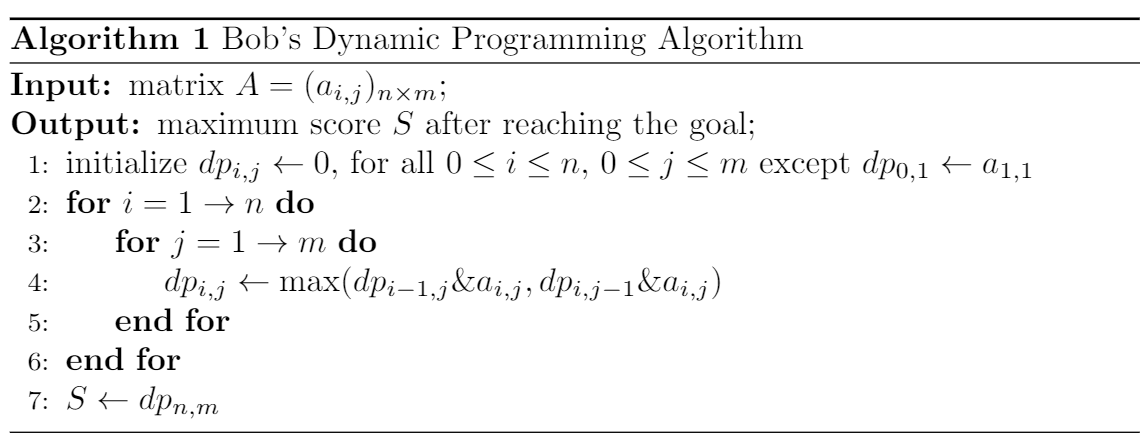}
    \begin{tcolorbox}[breakable, title=Gemini Vision Pro]
    The image shows Bob's dynamic programming algorithm. It takes a matrix A as input and outputs the maximum score S after reaching the goal. The algorithm initializes a 2D array dp with all zeros, except for \red{dp[0, 0]} which is set to A[1, 1]. Then, for each row i from 1 to n, the algorithm iterates through each column j from 1 to m and updates dp[i, j] to the maximum of dp[i - 1, j] \& A[i, j], dp[i, j - 1] \& \red{A[i, j], and dp[i - 1, j - 1] \& A[i, j]}. Finally, the algorithm returns dp[n, m].
    \end{tcolorbox}
    
    \begin{tcolorbox}[breakable, title=GPT-4V]
    The image shows a text box with the title "Algorithm 1 Bob's Dynamic Programming Algorithm" and contains pseudocode for a dynamic programming approach to solve a problem related to a game named "Walk on Matrix". The algorithm takes as input an n by m matrix A with elements $a_{i,j}$ and outputs the maximum score S after reaching the goal. The pseudocode outlines the steps of initializing a 2D array dp with zeros, except for dp[0,1] which is set to $a_{1,1}$. It then iterates through the matrix, updating the dp array with the maximum of two bitwise AND operations involving the current element and its adjacent elements to the left and above. The final score S is set to the value of dp[n,m].
    \end{tcolorbox}
\end{tcolorbox}
    \caption{An example of a \texttt{Pseudocode} image. Gemini Pro Vision makes a mistake in the array index and hallucinates, while GPT-4V's caption is correct.}
    \label{fig:appendix:caption:pseudocode}
\end{figure}

\begin{figure}[h!]
    \centering
\begin{tcolorbox}[title=2D Geometry]
    \includegraphics[width=\textwidth]{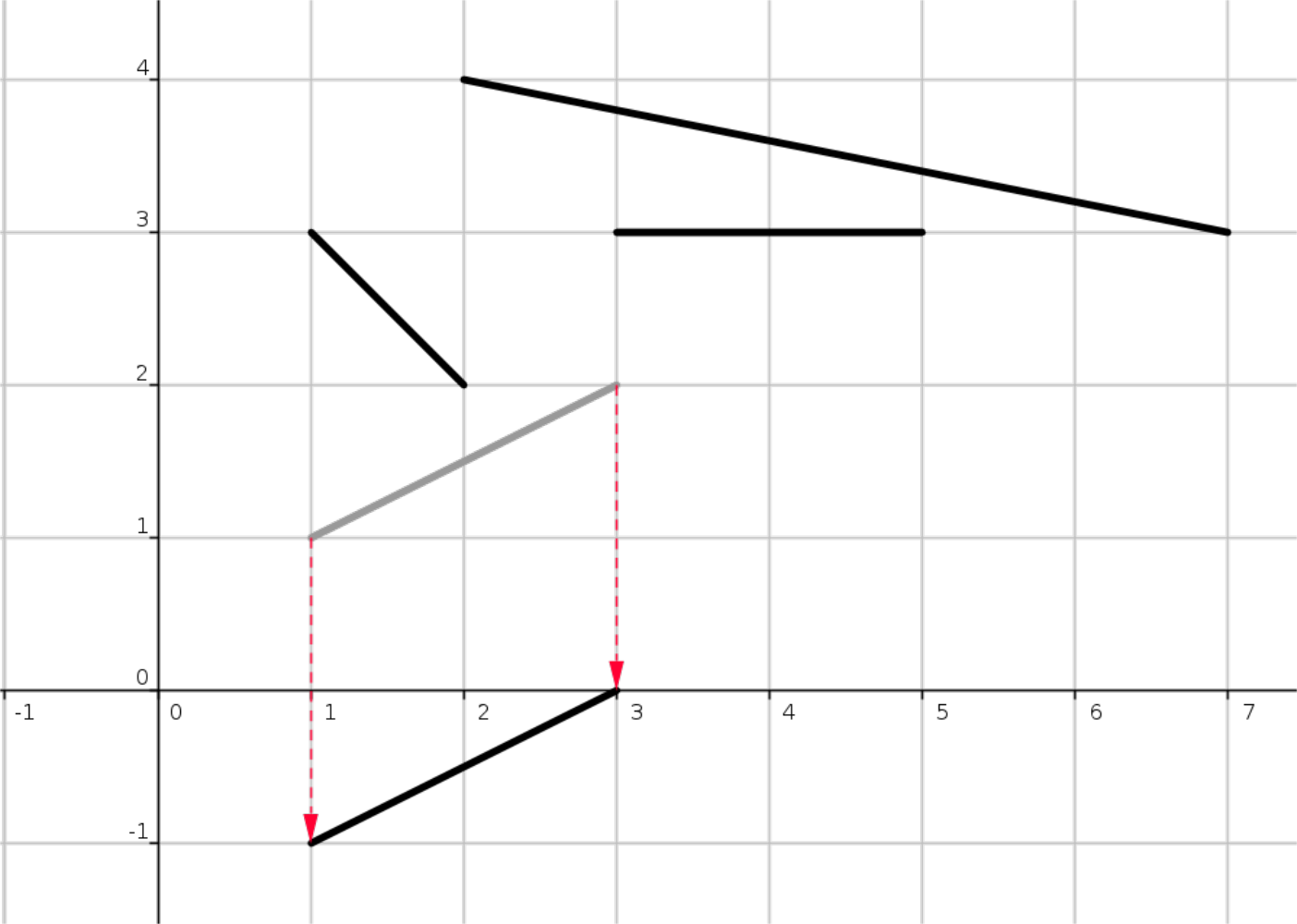}
    \begin{tcolorbox}[breakable, title=Gemini Vision Pro]
        The image shows a square representing the table with four sticks on it. The sticks are represented by line segments. The first stick is a diagonal line from (1, 3) to (2, 2). The second stick is a \red{horizontal} line from (1, 1) to (3, 2). The third stick is a diagonal line from (2, 4) to (7, 3). The fourth stick is a \red{vertical} line from (3, 3) to (5, 3). The red dotted lines show the shortest paths for removing the sticks. The first stick is removed by moving it down to the x-axis. \red{The second stick is removed by moving it left to the x-axis. The third stick is removed by moving it down to the x-axis. The fourth stick is removed by moving it left to the x-axis.}
    \end{tcolorbox}
    
    \begin{tcolorbox}[breakable, title=GPT-4V]
        The image shows a Cartesian coordinate system with a grid and three line segments representing sticks on a table. The x-axis is labeled from -1 to 7, and the y-axis is labeled from -1 to 4. There are three sticks: one is almost horizontal at y=3, another is slanted and \red{crosses the y-axis at y=1, and the third is vertical at x=3}. Red dashed arrows indicate the direction in which the sticks should be moved towards the x-axis. \red{The vertical stick} has an arrow pointing directly downward, while the slanted stick has an arrow pointing diagonally down to the nearest point on the x-axis. The horizontal stick does not have an arrow, but it is implied that it would move straight down as well. This illustrates the rule that sticks must be moved towards the x-axis along the shortest path without rotation or touching other sticks.
    \end{tcolorbox}

\end{tcolorbox}
    \caption{An example of a \texttt{2D Geometry} image. Both models struggle to discern the segments and extract their attributes.}
    \label{fig:appendix:caption:2d}
\end{figure}

% az_1334
\begin{figure}[h!]
    \centering
\begin{tcolorbox}[title=3D Geometry]
    \includegraphics[width=\textwidth]{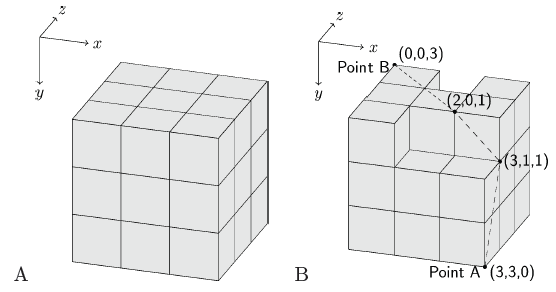}

    \begin{tcolorbox}[breakable, title=Gemini Vision Pro]
    Figure J.1A shows the largest colony, which consists of $3 \times 3 \times 3$ cubic blocks. Figure J.1B shows a smaller colony with some blocks missing. The dashed line in Figure J.1B is an example of \red{a} cable connecting \red{two} points on the surface of the colony.
        
    \end{tcolorbox}
    
    \begin{tcolorbox}[breakable, title=GPT-4V]
    The image contains two parts labeled as Figure J.1A and Figure J.1B, illustrating examples of space colonies made up of cubic blocks. Both parts show a three-dimensional coordinate system with axes labeled x, y, and z.\\ \\ Figure J.1A depicts a large colony that resembles a Rubik's cube, consisting of $3 \times 3 \times 3$ cubic blocks, forming a perfect cube.\\ \\Figure J.1B shows a smaller colony with some cubic blocks missing. The colony is irregularly shaped due to the missing blocks. There are dashed lines indicating the shortest path for a cable on the surface of the colony, connecting two points labeled as Point A (3,3,0) and Point B (0,0,3). The path of the cable wraps around the surface of the blocks, illustrating the requirement that the cable must remain on the surface.
        
    \end{tcolorbox}
\end{tcolorbox}

    \caption{An example of a \texttt{3D} image. Gemini Vision Pro only saw one of the segmented lines (called ``cables'' in the problem). GPT-4V interprets the image correctly.}
    \label{fig:appendix:caption:3d}
\end{figure}

% cf_1110_f
\begin{figure}[h!]
    \centering
\begin{tcolorbox}[title=Tree]
    \centering
    \includegraphics[width=0.3\textwidth]{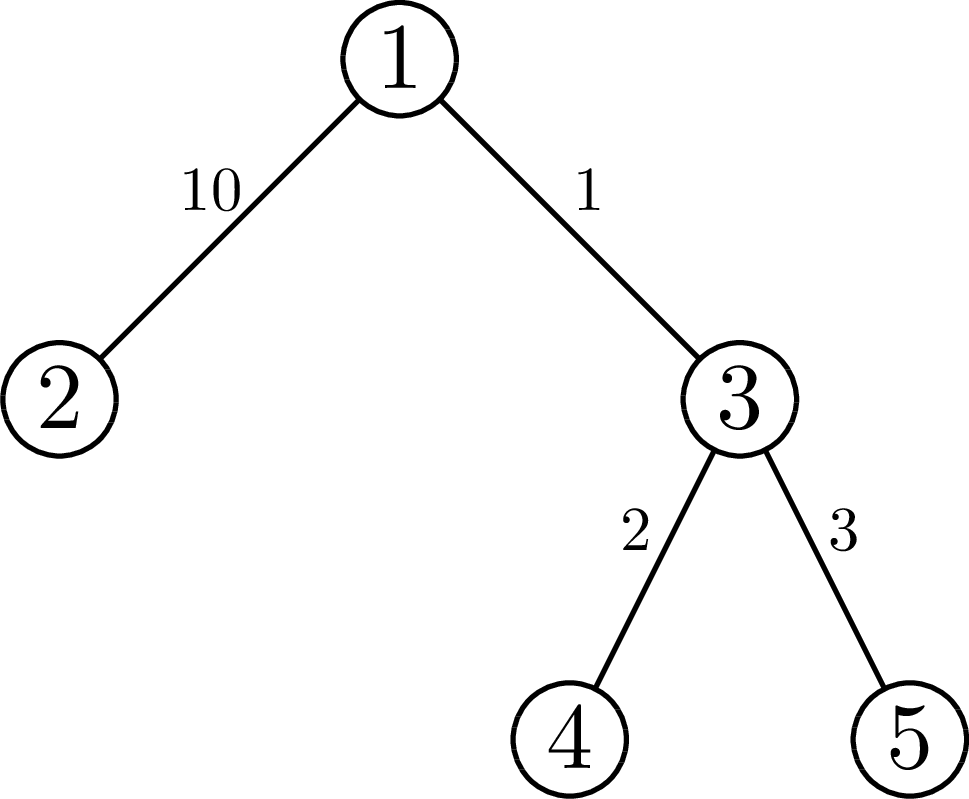}

    \begin{tcolorbox}[breakable, title=Gemini Vision Pro]
    The image shows a tree with 5 vertices and 4 edges. The vertices are numbered from 1 to 5. The edges are: (1, 2) with weight 10, (1, 3) with weight 1, (3, 4) with weight 2, and (3, 5) with weight 3.
        
    \end{tcolorbox}
    
    \begin{tcolorbox}[breakable, title=GPT-4V]
    The image shows a tree graph with 5 vertices labeled 1 through 5. Vertex 1 is connected to vertex 2 with an edge weight of 10 and to vertex 3 with an edge weight of 1. Vertex 3 is further connected to vertices 4 and 5 with edge weights of 2 and 3, respectively. The graph is drawn with vertices represented as circles containing their labels, and the edges are lines connecting these circles with the weights written alongside.
        
    \end{tcolorbox}
\end{tcolorbox}

    \caption{An example of a \texttt{Tree} image. Both models generate correct captions.}
    \label{fig:appendix:caption:tree}
\end{figure}

% cf_512_E
\begin{figure}[h!]
    \centering
\begin{tcolorbox}[title=Graph]
    \includegraphics[width=\textwidth]{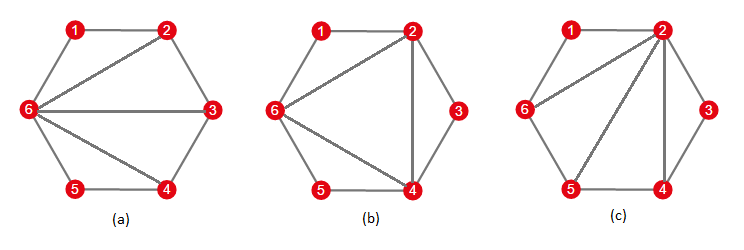}

    \begin{tcolorbox}[breakable, title=Gemini Vision Pro]
    The image shows \red{a} regular hexagon with 6 vertices labeled 1 through 6. There are 3 diagonals drawn in the hexagon, forming a triangulation.
        
    \end{tcolorbox}
    
    \begin{tcolorbox}[breakable, title=GPT-4V]
    The image shows three figures labeled (a), (b), and (c), each representing a different triangulation of a hexagon (a six-edged polygon). Each vertex of the hexagon is numbered from 1 to 6 in a clockwise direction. The figures illustrate the process of transforming one triangulation into another by flipping diagonals.\\ \\ Figure (a) shows the initial triangulation with diagonals connecting vertices \red{1-4, 1-5, and 5-3}.\\ Figure (b) shows an intermediate step after flipping the diagonal 6-3 to 2-4.\\ Figure (c) shows the goal triangulation with diagonals connecting vertices \red{1-4}, 2-4, and \red{2-3}.
        
    \end{tcolorbox}
\end{tcolorbox}

    \caption{An example of a \texttt{Graph} image. Gemini Vision Pro fails to recognize all three hexagons and the caption is vague, missing key information of how the hexagons are constructed. GPT-4V summarizes all subfigures but hallucinates on the vertices connected.}
    \label{fig:appendix:caption:graph}

\end{figure}

\begin{figure}[h!]
    \centering
\begin{tcolorbox}[title=Table]
\centering
    \includegraphics[width=0.35\textwidth]{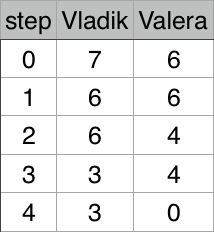}

    \begin{tcolorbox}[breakable, title=Gemini Vision Pro]
    The image shows a 2x3 table with the following values:\\
    \\
    step | Vladik | Valera\\
    -------|--------|--------\\
    0 | 7 | 6\\
    1 | 6 | \red{7}\\
    2 | \red{5} | \red{8}\\
    3 | \red{4} | \red{9}\\
    4 | 3 | \red{10}\\
    \red{5 | 2 | 11\\
    6 | 1 | 12\\
    7 | 0 | 13}

    \end{tcolorbox}
    
    \begin{tcolorbox}[breakable, title=GPT-4V]
    The image shows a table with three columns and \red{three} rows. The columns are labeled "step", "Vladik", and "Valera". The rows represent the steps of the candy exchange process. \red{At step 0, Vladik has 1 candy, and Valera also has 1 candy. At step 1, after Vladik gives 1 candy to Valera, Vladik has 0 candies left, and Valera has 2 candies.}
        
    \end{tcolorbox}
\end{tcolorbox}

    \caption{An example of a \texttt{Table}. Surprisingly, neither of the models were able to transcript the table correctly.} 
    \label{fig:appendix:caption:table}
\end{figure}

% ac_abc088_d
\begin{figure}[h!]
    \centering
\begin{tcolorbox}[title=Map]
    \includegraphics[width=\textwidth]{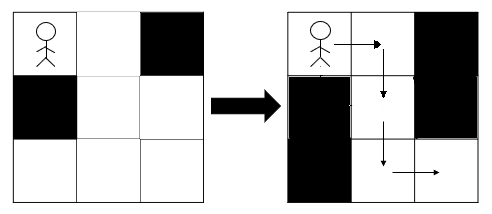}

    \begin{tcolorbox}[breakable, title=Gemini Vision Pro]
    The image shows a 3x3 grid with a character called Kenus at square (1, 1). The squares are painted black or white. Kenus can move up, down, left or right by one square. The game is completed when Kenus reaches square (3, 3) passing only white squares. Snuke can change the color of some of the white squares to black before the beginning of the game. However, he cannot change the color of square (1, 1) and (3, 3).        
    \end{tcolorbox}
    
    \begin{tcolorbox}[breakable, title=GPT-4V]
    The image shows a 3x3 grid with a stick figure representing Kenus at the top left corner (square 1,1). The grid has black and white squares, with the black squares represented by a solid fill and the white squares by an absence of fill. The left side of the image shows the initial state of the grid, with the top right square (1,3) and the middle left square (2,1) being black. The right side of the image shows the same grid after two white squares have been changed to black, specifically \red{the middle square (2,2) and the bottom middle square (3,2)}. Arrows indicate the path Kenus can take to reach the bottom right corner (3,3), moving right, down, down, and right, avoiding black squares.
        
    \end{tcolorbox}
\end{tcolorbox}

    \caption{An example of a \texttt{Map} image. Gemini Pro Vision's caption is correct overall, but important details about the path is missing. GPT-4V recognizes the switched squares, but grounds them to wrong positions.}
    \label{fig:appendix:caption:map}
\end{figure}

\begin{figure}[h!]
    \centering
\begin{tcolorbox}[title=Chessboard]
    \centering
    \includegraphics[width=0.3\textwidth]{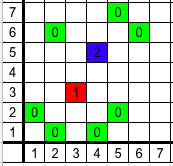}

    \begin{tcolorbox}[breakable, title=Gemini Vision Pro]
    The image shows a 7x7 chessboard with 7 green knights placed on it. The green knights are placed in the following cells: (1, 2), (2, 1), (4, 1), (5, 2), (5, 7), (6, 6), and \red{(7, 6)}. There is a red knight in cell (3, 3), and a blue knight in cell (4, 5).
        
    \end{tcolorbox}
    
    \begin{tcolorbox}[breakable, title=GPT-4V]
    The image shows a 7x7 section of an infinite chessboard with rows labeled 1 to 7 from bottom to top and columns labeled 1 to 7 from left to right. There are \red{five} cells marked with green zeroes, representing the initial knights' positions. These cells are located at coordinates (1,2), (2,1), (2,6), (5,2), and (5,7)\red{.} There is one cell marked with a red one at coordinates (3,3), indicating a knight placed during the process because it is under attack by at least four knights. Another cell is marked with a blue two at coordinates (4,5), showing that it will also have a knight placed in it after the knight in the red cell is added, as it will then be under attack by four knights. The rest of the cells are empty.
        
    \end{tcolorbox}
\end{tcolorbox}

    \caption{An example of a \texttt{Chessboard} image. The descriptions are mostly correct, but both models fail to list all green squares.}
    \label{fig:appendix:caption:chessboard}
\end{figure}

\begin{figure}[h!]
    \centering
\begin{tcolorbox}[title=Math]
    \centering
    \includegraphics[width=0.3\textwidth]{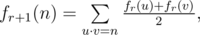}

    \begin{tcolorbox}[breakable, title=Gemini Vision Pro]
    The image shows a screenshot of a programming problem. The problem statement is as follows:

    \red{...(Repeats the problem statement)} $f_{r + 1}$ is defined as:
    $$f_{r + 1}(n) = \sum\limits_{\red{u^2 - v^2} = n} \frac{f_r(u) + f_r(v)}{2}$$
    \red{...(The rest of the problem statement)}
    \end{tcolorbox}
    
    \begin{tcolorbox}[breakable, title=GPT-4V]
    The image shows a mathematical function definition for $f_{r+1}(n)$ which is equal to the sum over all ordered pairs $(u, v)$ such that $u * v = n$ of the average of $f_r(u)$ and $f_r(v)$. The function $f_r$ is defined recursively, and the base function $f_0(n)$ counts the number of ways of factoring n into two co-prime factors p and q. The image is a part of a programming problem statement that involves calculating the value of $f_r(n)$ for given values of r and n, modulo $10^9 + 7$.        
    \end{tcolorbox}
\end{tcolorbox}

    \caption{An example of a \texttt{Math} image. Gemini Pro Vision's output contains unnecessary parts, and the subscript of the recognized equation is wrong. GPT-4V's caption is accurate.}
    \label{fig:appendix:caption:math}
\end{figure}

% \begin{figure}[h!]
%     \centering
% \begin{tcolorbox}[title=Math]
%     \includegraphics[width=\textwidth]{figures/sample_data/math.png}

%     \begin{tcolorbox}[breakable, title=Gemini Vision Pro]
%     The image shows a 3x3 matrix A with elements 0 and 1.
        
%     \end{tcolorbox}
    
%     \begin{tcolorbox}[breakable, title=GPT-4V]
%     The image shows a 3x3 matrix with parentheses around it, indicating that it is a single matrix. The matrix contains binary values (0s and 1s) arranged as follows:\\1 1 1\\0 1 1\\1 0 0
        
%     \end{tcolorbox}
% \end{tcolorbox}

%     \caption{An example of a \texttt{Pseudocode} image.}
%     \label{fig:appendix:caption:2d}
% \end{figure}

% cf_115_C
\begin{figure}[h!]
    \centering
\begin{tcolorbox}[title=Pattern]
    \centering
    \includegraphics[width=0.3\textwidth]{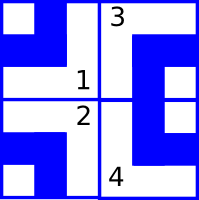}

    \begin{tcolorbox}[breakable, title=Gemini Vision Pro]
    The image shows a 2x2 grid with 4 pipe segments. The segments are numbered 1, 2, 3, and 4.\\ 
    \red{Segment 1 is a horizontal segment at the top of the grid.\\ 
    Segment 2 is a vertical segment at the left side of the grid.\\ 
    Segment 3 is a horizontal segment at the bottom of the grid.\\ 
    Segment 4 is a vertical segment at the right side of the grid.
    }
        
    \end{tcolorbox}
    
    \begin{tcolorbox}[breakable, title=GPT-4V]
    The image shows four types of pipe segments, each represented by a square with two arrows indicating the ends of the pipe. Segment 1 has ends at the top and left, segment 2 has ends at the bottom and left, segment 3 has ends at the \red{top} and right, and segment 4 has ends at the \red{bottom} and right.
        
    \end{tcolorbox}
\end{tcolorbox}

    \caption{An example of a \texttt{Pattern} image. GPT-4V succeeds in describing two of the four blocks, whereas Gemini Vision Pro fails to interpret any.}
    \label{fig:appendix:caption:pattern}
\end{figure}

\begin{figure}[h!]
    \centering
\begin{tcolorbox}[title=Others]
    \centering
    \includegraphics[width=0.75\textwidth]{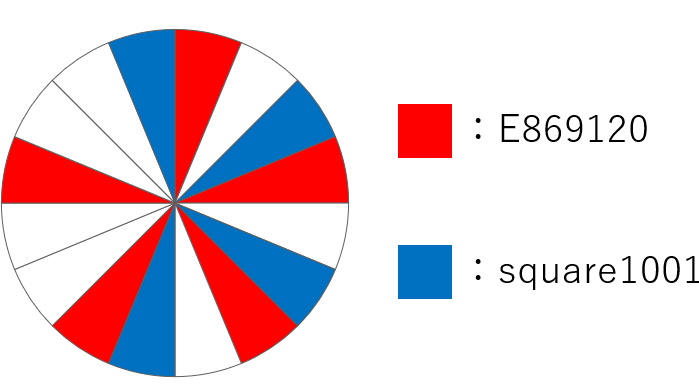}

    \begin{tcolorbox}[breakable, title=Gemini Vision Pro]
    The image shows a round cake cut into 16 equal fan-shaped pieces.\\ The pieces are arranged in a circle, and each piece is labeled with a number from 1 to 16.\\ The colors of the pieces alternate between red and blue.
        
    \end{tcolorbox}
    
    \begin{tcolorbox}[breakable, title=GPT-4V]
    The image shows a round cake cut into 16 equal fan-shaped pieces. The pieces are alternately colored in red and blue to represent the pieces taken by E869120 and square1001, respectively. There are 5 red pieces and 4 blue pieces, with no two adjacent pieces of the same color, indicating that the instruction in the note has been followed.
        
    \end{tcolorbox}
\end{tcolorbox}

    \caption{An example of a \texttt{Others} image. Both models are correct.}
    \label{fig:appendix:caption:others}
\end{figure}

% \begin{figure}[h!]
%     \centering
% \begin{tcolorbox}[title=]
%     \includegraphics[width=\textwidth]{}

%     \begin{tcolorbox}[breakable, title=Gemini Vision Pro]

%     \end{tcolorbox}
    
%     \begin{tcolorbox}[breakable, title=GPT-4V]

%     \end{tcolorbox}
% \end{tcolorbox}

%     \caption{An example of a \texttt{Pseudocode} image.}
%     \label{fig:appendix:caption:2d}
% \end{figure}

%% file: sections/appendices/code_case_study.tex
\clearpage
\newpage
\section{Case Study: Code Analysis}
\label{appendix:sec:code_case_study}
In this section, some samples of machine-generated solution code are presented and studied. Incorrect segments are colored in red and bolded in the code listed.

\subsection{Incorrect and Inefficient solution}
\label{appendix:sec:code_case_study:tle_wa}
\begin{figure}[h!]
\begin{tcolorbox}[title=Problem]
    Let \(J(n)\) be a three-dimensional body that:
\begin{itemize}
    \item is a union of unit cubes whose all vertices lie on integer coordinates,
    \item contains all points that are closer than the distance of \(\sqrt{n}\) to the origin, and
    \item is the smallest of all such bodies.
\end{itemize}

Figure 1 below shows how \(J(1)\), \(J(2)\), and \(J(3)\) look.

\begin{center}
\includegraphics[scale=0.6]{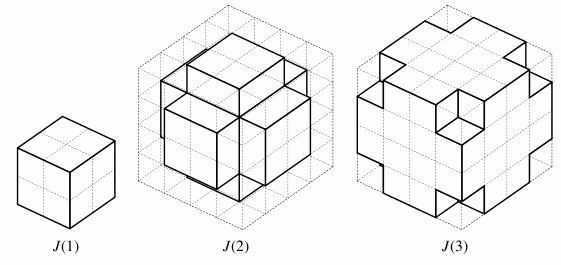}
\end{center}

Your task is to calculate how many faces \(J(n)\) have. Here, we define two squares belong to the same face if they are parallel and share an edge, but don’t if they share just a vertex.

\textbf{Input}

The input consists of multiple data sets, each of which comes with a single line containing an integer \(n\) (1 \(\leq n \leq 1000000\)). The end of input is indicated by \(n = 0\).

\textbf{Output}

For each data set, print the number of faces \(J(n)\) have.

\textbf{Sample Input}

\begin{verbatim}
1
2
3
4
0
\end{verbatim}

\textbf{Output for the Sample Input}

\begin{verbatim}
6
30
30
6
\end{verbatim}
\end{tcolorbox}

\caption{The description of the problem of the case study.}
\label{appendix:case_study_code:indfficiency_problem}
\end{figure}

Sometimes GPT-4V can generate inefficient code that takes too long to finish execution, leading to a time-out. An demonstration is given in Figure \ref{appendix:case_study_code:indfficiency_problem} and Figure \ref{appendix:case_study_code:inefficiency_code}. Moreover, it implements a wrong method of calculating the required quantity.

The problem statement is listed in Figure \ref{appendix:case_study_code:indfficiency_problem}. It asks to find the number of faces in a minimum 3D shape made of unit cubes that cover all integer coordinates within a sphere of radius $\sqrt{n}$. In Figure \ref{appendix:case_study_code:inefficiency_code}, GPT-4V's solution loops over all integer points inside $-\sqrt{n} \leq x, y, z \leq \sqrt{n}$. A cleverer method is to leverage the symmetry and only count the faces perpendicular to an axis, reducing the operations to $\frac{1}{6}$ of the original. Moreover, the algorithm for calculating the connected faces in the code is incorrect.

\begin{figure}
\begin{tcolorbox}[left=8mm]
\begin{minted}[linenos, numbersep=8pt, breaklines, escapeinside=||]{python}
import math

def count_faces(n):
    # Calculate the radius of the sphere
    r = math.sqrt(n)
    
    # Calculate the range for x, y, z
    limit = math.ceil(r)
    
    # Initialize the count of faces
    faces = 0
    
    # Iterate through all possible cubes and count the faces
    for x in range(|\red{\textbf{-limit, limit + 1}}|):
        for y in range(|\red{\textbf{-limit, limit + 1}}|):
            for z in range(|\red{\textbf{-limit, limit + 1}}|):
                # Check if the cube is on the surface of J(n)
                if r - 1 < math.sqrt(x**2 + y**2 + z**2) <= r:
                    # Check each face of the cube
                    |\red{\textbf{for dx, dy, dz in [(1, 0, 0), (0, 1, 0), (0, 0, 1)]:}}|
                        # If the neighboring cube is not part of J(n), count the face
                        |\red{\textbf{if math.sqrt((x+dx)**2 + (y+dy)**2 + (z+dz)**2) > r:}}|
                            |\red{\textbf{faces += 1}}|

    return faces

# Read input and process each case
while True:
    n = int(input())
    if n == 0:
        break
    print(count_faces(n))

\end{minted}
\end{tcolorbox}
\caption{An inefficient and incorrect solution generated by GPT-4V. }
\label{appendix:case_study_code:inefficiency_code}
\end{figure}

\subsection{TypeError caused by naming conflict}
\label{appendix:sec:code_case_study:name_conflict}
% CF_182_C

As mentioned in Section \ref{sec:case_study:code}, despite the advance in understanding complex questions and generating syntactically correct solutions, GPT-4V still makes basic errors. 
In this example code shown in Figure \ref{appendix:case_study_code:wrong_naming_code}, GPT-4V assigns the name \verb|len| to a local variable, which conflicts with and overrides the Python built-in \verb|len()| function. Consequently, when the code attempts to obtain the length of a list on line 14, it raises a \verb|TypeError|. This is because \verb|len| now refers to the variable, not the function it originally intended to use.

\renewcommand{\theFancyVerbLine}{\sffamily\textcolor{gray}{\small\arabic{FancyVerbLine}}}

\begin{figure}

\begin{tcolorbox}[left=8mm, title=GPT-4V's solution]

\begin{minted}[linenos, numbersep=8pt, breaklines, escapeinside=||]{python}
def calculate_optimal_sum(n, |\red{\textbf{len}}|, a, k):
    # Helper function to calculate the sum of a subarray
    def subarray_sum(i):
        return sum(a[i:i+len])
 
    # Calculate the initial optimal sum without any operations
    optimal_sum = max(subarray_sum(i) for i in range(n - len + 1))
 
    # Sort the array elements by their absolute values in descending order
    # This will help us to choose which elements to flip to maximize the sum
    negative_elements = sorted((x for x in a if x < 0), key=abs, reverse=True)
 
    # Perform up to k operations, flipping the most negative elements
    for i in range(min(k, |\red{\textbf{len}}|(negative_elements))):
        # Flip the negative element to positive
        negative_elements[i] *= -1
        # Update the array with the flipped element
        a[a.index(negative_elements[i] * -1)] = negative_elements[i]
 
    # Calculate the new optimal sum after performing the operations
    optimal_sum = max(optimal_sum, max(subarray_sum(i) for i in range(n - len + 1)))
 
    return optimal_sum
 
# Read input
n, |\red{\textbf{len}}| = map(int, input().split())
a = list(map(int, input().split()))
k = int(input())
 
# Calculate and print the result
print(calculate_optimal_sum(n, \textbf{len}, a, k))

\end{minted}

\end{tcolorbox}

\caption{A sample of GPT-4V's wrong solution caused by a name conflict.}
\label{appendix:case_study_code:wrong_naming_code}
\end{figure}

\clearpage
\subsection{Input Parsing Error}
\label{appendix:sec:code_case_study:input_parsing}
\begin{figure}[h!]
\begin{tcolorbox}[title=Problem]
A snowflake of order $n$ is formed by overlaying an equilateral triangle (rotated by $180$ degrees) onto each equilateral triangle of the same size in a snowflake of order $n-1$. A snowflake of order $1$ is a single equilateral triangle.
\begin{center}
\includegraphics[scale=0.6]{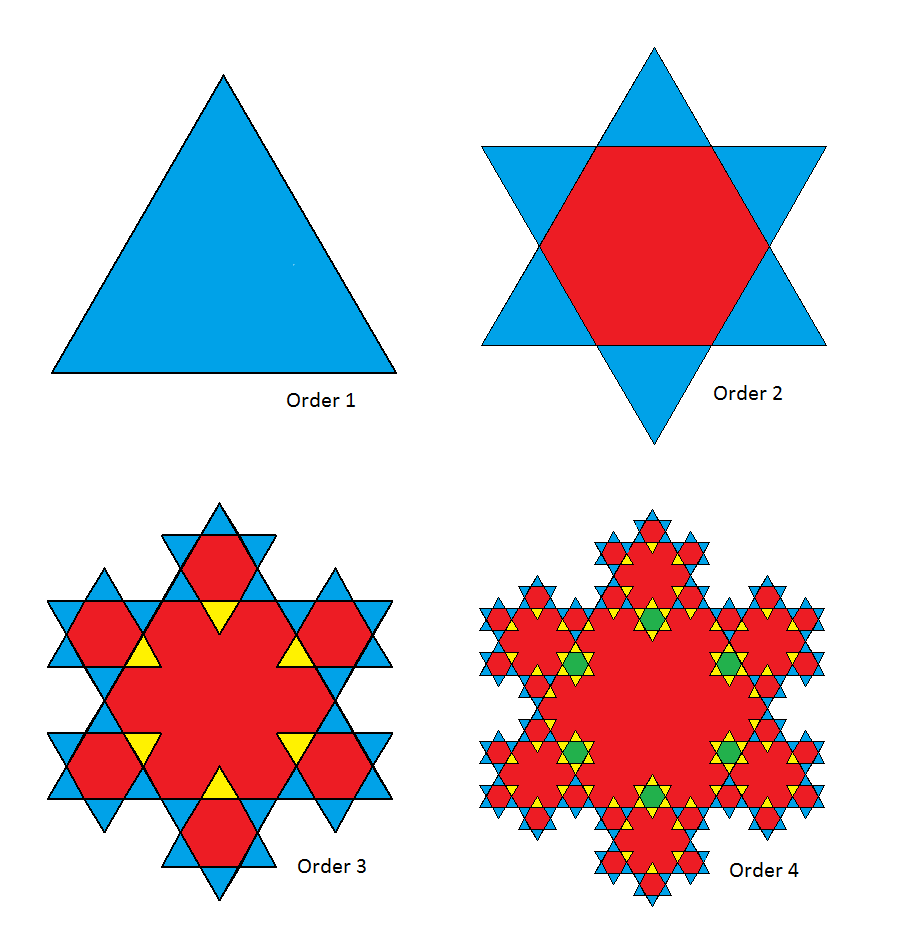}
\end{center}
Some areas of the snowflake are overlaid repeatedly. In the above picture, blue represents the areas that are one layer thick, red two layers thick, yellow three layers thick, and so on.

For an order $n$ snowflake, let $A(n)$ be the number of triangles that are one layer thick, and let $B(n)$ be the number of triangles that are three layers thick. Define $G(n) = \gcd(A(n), B(n))$.

E.g. $A(3) = 30$, $B(3) = 6$, $G(3)=6$.

$A(11) = 3027630$, $B(11) = 19862070$, $G(11) = 30$.

Further, $G(500) = 186$ and $\sum_{n=3}^{500}G(n)=5124$.

Find $\displaystyle \sum_{n=3}^{10^7}G(n)$.

\end{tcolorbox}

\caption{The description of the problem that causes an input parsing error.}
\label{appendix:case_study_problem:input_parsing}
\end{figure}

The problem shown in Figure \ref{appendix:case_study_problem:input_parsing} asks the summation of a certain function with input from $3$ to $10^7$. However, GPT-4V tries to read the upper limit from user input in line 24 in Figure \ref{appendix:case_study_code:input_parsing}, raising an EOFError (End of File Error) because the input is empty. 
% pe_570

\begin{figure}
\begin{tcolorbox}[left=8mm]
\begin{minted}[linenos, numbersep=8pt, breaklines, escapeinside=||]{python}
from math import gcd

def calculate_snowflake_layers(n):
    # Base values for A(3) and B(3)
    A = 30
    B = 6
    G = gcd(A, B)
    
    # Calculate A(n) and B(n) using the recursive relationship
    for i in range(4, n + 1):
        A = A * 3 + 12 * (i - 2)
        B = B * 3 + 6 * (i - 2)
        G = gcd(A, B)
    
    return G

def sum_of_gcds(n):
    total_sum = 0
    for i in range(3, n + 1):
        total_sum += calculate_snowflake_layers(i)
    return total_sum

# Read the upper limit from standard input
|\red{\textbf{upper\_limit = int(input())}}|

# Calculate and print the sum of GCDs
print(sum_of_gcds(upper_limit))
\end{minted}
\end{tcolorbox}
\caption{GPT-4V tries to read the upper limit from user input but it should be hardcoded $10^7$.}
\label{appendix:case_study_code:input_parsing}
\end{figure}